\documentclass[twoside,compsoc]{IEEEtran}
\usepackage{makecell}
\usepackage{hyperref}
\usepackage{array}
\usepackage{graphicx,amssymb,amsmath}
\usepackage{multicol}
\usepackage[noadjust]{cite}
\usepackage{setspace}
\usepackage{subfigure}
\usepackage{float}
\usepackage {url}
\usepackage{stfloats}
\usepackage{amsthm,pifont}
\usepackage{flushend}
\usepackage{cases,subeqnarray}
\usepackage{bm,multirow,bigstrut}
\usepackage{amsmath, amsthm, amssymb}
\usepackage{textcomp}
\usepackage{latexsym,bm}
\usepackage{booktabs}
\usepackage{xcolor}
\usepackage{mathtools}
\usepackage{dsfont}
\usepackage{extarrows}
\usepackage{epsfig}
\usepackage{epsfig}
\usepackage{epstopdf}
\usepackage{colortbl}
\usepackage{algpseudocode}
\usepackage{algorithmicx,algorithm}

\theoremstyle{plain}

\theoremstyle{plain}

\usepackage{amsmath}

\definecolor{Gray}{gray}{0.85}
\IEEEoverridecommandlockouts
\begin{document}
\title{{\textit{ReaCritic}}: Reasoning Transformer-based DRL Critic-model Scaling For Wireless Networks}

\author{Feiran You, Hongyang Du
\thanks{F. You and H. Du are with the Department of Electrical and Electronic Engineering, University of Hong Kong, Pok Fu Lam, Hong Kong SAR, China (email: fryou@eee.hku.hk, duhy@eee.hku.hk).}
}
\maketitle
\vspace{-1cm}

\begin{abstract}
Heterogeneous Networks (HetNets) pose critical challenges for intelligent management due to the diverse user requirements and time-varying wireless conditions. These factors introduce significant decision complexity, which limits the adaptability of existing Deep Reinforcement Learning (DRL) methods. In many DRL algorithms, especially those involving value-based or actor-critic structures, the critic component plays a key role in guiding policy learning by estimating value functions. However, conventional critic models often use shallow architectures that map observations directly to scalar estimates, limiting their ability to handle multi-task complexity. In contrast, recent progress in inference-time scaling of Large Language Models (LLMs) has shown that generating intermediate reasoning steps can significantly improve decision quality. Motivated by this, we propose {\textit{ReaCritic}}, a reasoning transformer-based critic-model scaling scheme that brings reasoning-like ability into DRL. {\textit{ReaCritic}} performs horizontal reasoning over parallel state-action inputs and vertical reasoning through deep transformer stacks. It is compatible with a broad range of value-based and actor-critic DRL algorithms and enhances generalization in dynamic wireless environments. Extensive experiments demonstrate that \textit{ReaCritic} improves convergence speed and final performance across various HetNet settings and standard OpenAI Gym control tasks. {\color{black}The code of \textit{ReaCritic} is available at https://github.com/NICE-HKU/ReaCritic.}
\end{abstract}
\IEEEpeerreviewmaketitle
\section{Introduction}
    The rapid growth of intelligent wireless services such as the Internet of Things (IoT), Mobile Edge Computing (MEC), and smart city applications is driving the evolution of future wireless networks toward large-scale Heterogeneous Networks (HetNets)~\cite{tian2024large}. These HetNets involve a large number of mobile users and dynamic environments with heterogeneous communication and computing resources~\cite{pivoto2025comprehensive}. \textcolor{black}{The efficient management of these infrastructures faces three primary technical bottlenecks that hinder traditional resource management. First, the massive scale of mobile users and multi-tier base stations results in an explosive growth of state-action dimensionality whereby the search for optimal configurations becomes computationally prohibitive for conventional optimization techniques~\cite{gu2025task}. Second, the coexistence of heterogeneous network tiers creates intricate spatial-temporal couplings in which resource allocation at a specific access point exerts complex non-linear interference on neighboring nodes~\cite{lu2026agentic}, thereby complicating the derivation of a global optimum in multi-user interference regimes. Third, the intrinsic non-stationarity of the wireless medium, driven by rapid user mobility, stochastic channel fading, and unpredictable traffic bursts, necessitates a decision-making framework capable of robust real-time adaptation rather than relying on static or predefined models~\cite{ning2025survey}.}

\textcolor{black}{As a result, HetNets must handle massive data traffic, ensure low-latency responses, and maintain energy efficiency under diverse service demands, which introduces fundamental challenges for scalable resource management and stable decision-making~\cite{agarwal2022comprehensive}. Specifically, the multifaceted requirements for diverse Quality-of-Service (QoS) levels, combined with the time-varying nature of wireless conditions, result in high-dimensional state spaces and complex user-resource interactions. Such properties fundamentally limit the scalability and generalization of traditional optimization and supervised learning-based methods, as they often fail to capture the latent dependencies required to maintain performance in non-stationary environments~\cite{zhou2024large}.}
\begin{figure}[t]
\centering
\includegraphics[width = 0.49\textwidth]{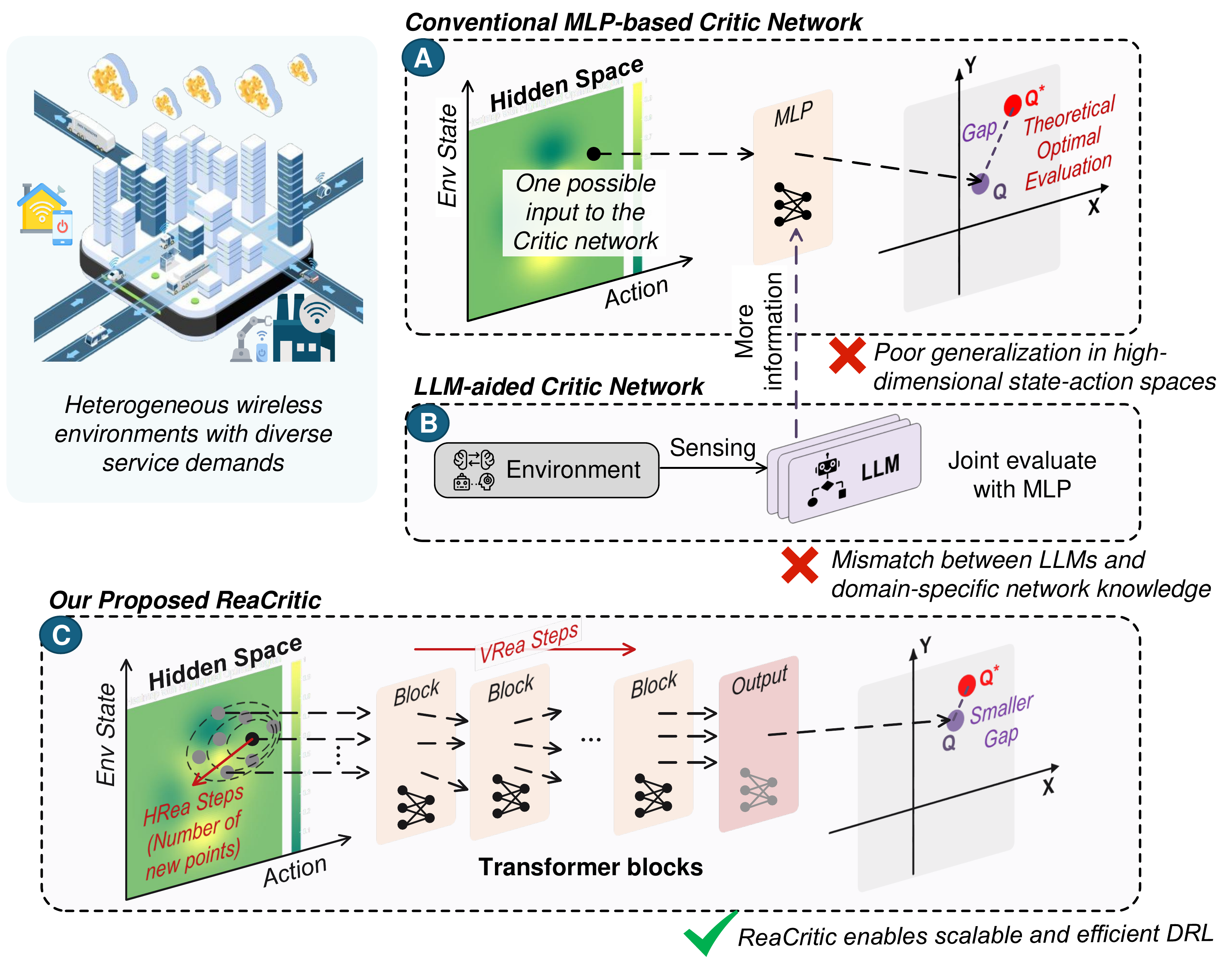}
\vspace{-0.2cm}
\caption{Comparison of DRL critic designs for HetNets. Conventional MLP-based critics (Part A) struggle with generalization in high-dimensional spaces. LLM-aided critics (Part B) improve expressiveness but face domain mismatch and integration challenges. Our proposed ReaCritic (Part C) introduces transformer-based bidirectional reasoning to achieve more accurate value estimation and scalable DRL training.}
\label{fig_mo}
\end{figure}
Faced with these challenges, Deep Reinforcement Learning (DRL) has gained attention as a data-driven approach for autonomous network management in complex wireless environments~\cite{sun2025comprehensive}. By interacting with the environment and optimizing long-term rewards, DRL enables agents to adapt to dynamic network conditions without relying on predefined models~\cite{alwarafy2022frontiers}.
Its success in domains such as intelligent control, robotics, and wireless resource management demonstrates its potential for handling large-scale, uncertain systems~\cite{du2024enhancing,won2024resource,yang2024beyond}. 
However, when applied to large-scale HetNets, DRL still faces key limitations not only due to environmental complexity but also due to internal architectural bottlenecks. 
Although DRL algorithms are capable of learning adaptive behaviors through trial-and-error interaction, many existing designs rely on simple Multi-Layer Perceptron (MLP)-based model structures that treat input states as flat feature vectors and apply direct approximations for policy or value functions~\cite{vasudevan2024machine}. 
This formulation fundamentally lacks the capacity to capture latent dependencies among state variables or to represent tradeoffs inherent in multi-objective decision-making. 
In heterogeneous wireless environments, where metrics such as latency, throughput, and energy efficiency are often tightly coupled and evolve asynchronously, such limitations can lead to unstable training, slow convergence, and poor performance in dynamic or unseen scenarios~\cite{hou2025distributed,he2024age}. Taking actor-critic-based DRL algorithms as an example, as shown in Fig.~\ref{fig_mo}, the critic network estimates Q-values based on environmental states and the actions selected by the actor network.However, the high dimensionality and structural complexity of HetNet states and the action space make this mapping difficult and inaccurate, limiting actor updates and ultimately degrading overall DRL performance.

Therefore, it is essential to enhance the scalability of DRL frameworks to better accommodate high-dimensional state spaces and dynamic network behaviors. One common strategy is to increase model depth or width, with the goal of improving the DRL's capacity to interpret complex input features. However, scaling up the MLP parameter size alone often leads to overfitting, unstable convergence, and limited robustness in non-stationary environments. A more effective direction is to improve the DRL model through architectural adjustments that enable it to process structured observations and capture latent dependencies within the environment, beyond what can be achieved through conventional model size expansion. 
In this context, the architectural principles of Large Language Models (LLMs) offer a compelling source of inspiration. Recent advances in reasoning-capable LLMs have demonstrated strong capabilities in pattern abstraction and generalization across a wide range of tasks~\cite{hagos2024recent}. These LLMs are effective at modeling intricate relationships and long-term dependencies in structured or semi-structured data, opening opportunities for improving representation learning and decision-making in complex environments. 

Motivated by the capabilities of LLMs, recent research has explored LLM-aided DRL frameworks, where well-pretrained LLMs are employed to assist in policy generation, reward shaping, or action refinement~\cite{cao2024survey,sun2025generative,he2024large,du2024mixture}. While these approaches have shown promise in certain domains, directly incorporating LLMs into the DRL training loop introduces several critical limitations. 
First, LLMs are primarily designed for natural language understanding and generation, which does not naturally align with the optimization-driven, reward-sensitive objectives of DRL. Second, LLMs exhibit non-deterministic behavior that is often difficult to control, resulting in unstable learning dynamics and inconsistent policy updates~\cite{huang2025look}. Third, integrating large-scale LLMs incurs substantial computational overhead and lacks theoretical support regarding convergence and stability in control settings~\cite{wu2024evolutionary}. These limitations indicate that enhancing the representation and reasoning capacity of DRL through architectural design remains a key challenge:
\begin{itemize}
\item \textbf{Q1 (How):} How can LLMs’ reasoning capability be effectively adapted to DRL settings without directly using pre-trained LLMs as external modules?
\item \textbf{Q2 (Where):} Which DRL component is most suitable for integrating LLM-inspired reasoning and scaling to support complex, dynamic environments?
\end{itemize}

To address the above questions, it is necessary to first reflect on the source of LLMs’ reasoning capabilities. Fundamentally, these capabilities do not arise from language data alone, but from the architectural properties of Transformer networks. Through self-attention and hierarchical composition, Transformers can capture relational dependencies and long-range context, forming the backbone of scalable reasoning in LLMs. This insight suggests that one can extract and adapt the structural mechanisms underlying LLMs' reasoning processes. Recent works have explored Transformer-based designs to improve value estimation, policy modeling, and trajectory learning in complex environments~\cite{li2022lane,wang2021deep}. 
For instance, the authors in~\cite{li2022lane} proposed a lightweight Transformer model for autonomous driving that combines temporal semantic extraction and risk modeling to generate safer driving strategies. Similarly, the authors in~\cite{wang2021deep} developed SACCT, a DRL framework for adaptive live streaming, where a communication Transformer captures multi-scale dependencies to jointly optimize uplink scheduling and edge transcoding. 
However, existing approaches primarily focus on task-specific modeling enhancements and do not explicitly introduce structured reasoning processes that support scalable, general-purpose decision-making across dynamic network environments.

To address {\textbf{Q1 (How)}}, \textcolor{black}{we adopt a Transformer-based reasoning architecture as a more stable and scalable alternative to conventional designs. Unlike traditional models, ReaCritic addresses HetNet challenges through two synergistic reasoning dimensions. First, the Horizontal Reasoning (HRea) module generates parallel reasoning tokens to explore a broader range of state-action correlations, effectively managing the breadth of high-dimensional input features. Second, the Vertical Reasoning (VRea) module utilizes stacked abstraction layers to perform hierarchical logic distillation, providing the depth needed to achieve stable convergence in non-stationary environments.} To address {\textbf{Q2 (Where)}}, we embed this reasoning-like capability into the critic module of the DRL framework. As the critic is responsible for value estimation and provides gradient signals to guide policy updates, it is the natural component for incorporating structured inference. \textcolor{black}{By systematically scaling these reasoning dimensions, ReaCritic enables the critic to adapt its computational effort to the complexity of the task, thereby ensuring superior performance and robustness in large-scale wireless communication scenarios.} This choice preserves the original policy structure while substantially improving the agent’s ability to model long-term dependencies, reason over complex state-action trajectories, and adapt to heterogeneous and time-varying network environments. 
Our proposed design, referred to as \textit{ReaCritic}, is illustrated in Part C of Fig.~\ref{fig_mo}. The algorithm can be centrally deployed, which avoids excessive computational load on edge devices and enables scalable coordination across heterogeneous users. In highly coupled environments, such centralized operation ensures globally consistent decisions and remains tractable for large, dynamic networks where performance is not heavily constrained by real-time latency.

The main contributions of this paper are as follows:

\begin{itemize}
\item We formulate a DRL-based resource management problem for large-scale HetNets, capturing the joint effects of heterogeneous user types, time-varying wireless conditions, and asynchronous service demands. This formulation reflects key challenges such as non-stationarity and high dimensionality, motivating the need for scalable DRL frameworks with enhanced reasoning capabilities.
\item We propose {\textit{ReaCritic}}, a reasoning transformer-based critic-model scaling architecture for DRL, inspired by inference-time scaling in LLMs. {\textit{ReaCritic}} enables horizontal reasoning across parallel state-action token sequences and vertical reasoning through stacked transformer blocks, supporting adaptable Q value estimation in complex environments.
\item We conduct extensive experiments in a large-scale multi-user HetNet environment to evaluate {\textit{ReaCritic}}. Compared with MLP-based critics and standard DRL baselines, {\textit{ReaCritic}} achieves faster convergence, improved policy stability, and stronger adaptability under dynamic and wireless network conditions.
\end{itemize}

The remainder of this paper is organized as follows: Section~\ref{related} discusses related work in HetNets, DRL for networks, and transformer for wireless
networks. Section~\ref{system} presents the wireless network system model, and gives the problem formulation of the optimization problem in the considered HetNet. Section~\ref{large} details the design of {\textit{ReaCritic}} and the reasoning transformer-based critic-model scaling. Section~\ref{num} presents comprehensive experimental results and analysis. Finally, Section~\ref{conclusion} concludes the paper with a summary of our key findings.
A list of mathematical symbols frequently used in this paper is shown in Table~\ref{table1xxx}.
\begin{table}[t]
\caption{Mathematical Notations}
\vspace{-0.2cm}
\centering
\label{table1xxx}
\renewcommand{\arraystretch}{1}
\begin{tabular}{m{1.8cm}|m{6.0cm}}
\toprule
\textbf{Notation} & \textbf{Description} \\
\midrule
$x_m$ & Position of the $m_{\rm th}$ user \\
$d_m$ & Computation demand (in CPU cycles) \\
$f_m$ & Local CPU frequency of the $m_{\rm th}$ user \\
$p_{\mathrm{u},m}, p_{\mathrm{d},m}$ & Normalized uplink/downlink power allocation factors \\
$b_{\mathrm{u},m}, b_{\mathrm{d},m}$ & Normalized uplink/downlink bandwidth allocation factors \\
$c_m$ & Normalized computation resource allocation factor\\
$q_{\mathrm{u},m}, q_{\mathrm{d},m}$ & Uplink/downlink Rician fading gain \\
$\sigma_{\mathrm{u}}^2, \sigma_{\mathrm{d}}^2$ & Uplink/downlink noise power \\
$P_{\mathrm{u},m}$ & Maximum uplink transmit power of the $m_{\rm th}$ user \\
$P_{\mathrm{d},\max}$ & Maximum downlink transmit power of the BS \\
$B_{\mathrm{u},m}, B_{\mathrm{d},m}$ & Maximum uplink/downlink bandwidth \\
$\zeta_m$ & Large-scale path loss of the $m_{\rm th}$ user \\
$C_{\max}$ & Total available computation capacity \\
$\rho_m$ & Local compute efficiency of the $m_{\rm th}$ user \\
$\tau_m$ & User type index \\
$Q_m$ & Utility function of the $m_{\rm th}$ user \\
$\alpha_m^{(i)}$ & Weight for the $i_{\rm th}$ objective \\
$\mathcal{R}_{\mathrm{u},m}, \mathcal{R}_{\mathrm{d},m}$ & Uplink/downlink data rate \\
$\eta_{\mathrm{u},m}, \eta_{\mathrm{d},m}$ & Uplink/downlink energy efficiency \\
$L_m$ & Service latency \\
$l_{\mathrm{th},m}$ & Latency threshold for the $m_{\rm th}$ user \\
$B$ & Batch size of DRL \\
$d_{\rm h}$ & The hidden dimension size\\
$H$ & The number of horizontal reasoning steps \\
$V$ & The number of vertical reasoning steps, i.e., vertical Transformer blocks\\
\bottomrule
\end{tabular}
\end{table}

\section{Related Work}\label{related}
In this section, we discuss related works, including HetNets, DRL, and transformer for wireless networks.

\subsection{ Heterogeneous Networks}
HetNets have been studied using a range of techniques, including optimization~\cite{xu2021survey}, game theory~\cite{gonzalez2024network}, heuristics~\cite{lotfolahi2023multi}, and more recently, DRL~\cite{li2022applications}. 
These approaches aim to address the diverse user demands and dynamic resource constraints inherent in HetNets. Despite progress, the high dimensionality and coupling among decision variables make HetNets a particularly challenging yet critical domain for intelligent network management~\cite{tezergil2022wireless}.
In~\cite{sharma2023secboost}, the authors proposed a secrecy-aware, energy-efficient scheme for HetNets, jointly optimizing power control, channel allocation, and beamforming under secrecy and Signal-to-Interference-plus-Noise Ratio (SINR) constraints. They formulated the problem as a Markov Decision Process (MDP) and developed a multi-agent DRL framework, i.e., SecBoost, using dueling Double Deep Q-Network (DDQN) and prioritized experience replay to maximize long-term secrecy energy efficiency.
In~\cite{ding2024drl}, a joint task offloading and resource management problem was formulated to improve computation efficiency in MEC-enabled HetNets. The authors addressed the resulting MINLP using an Advantage Actor-Critic (A2C)-based algorithm (A2C-JTRA), which jointly optimizes offloading decisions, transmit power, and CPU frequency for MEC servers and end devices.
In~\cite{sharma2023secrecy}, a secrecy-oriented joint optimization of beamforming, artificial noise, and power control was proposed for Tera Hertz (THz)-enabled femtocells. The authors used a Multi-Agent Reinforcement Learning (MARL) framework and introduced two DRL-based methods, i.e., Deep Deterministic Policy Gradient-based Secrecy Maximization (DDPG-SM) and Asynchronous-Advantage-Actor-Critic based Secrecy Maximization (A3C-SM), to operate in continuous action spaces and enhance long-term secrecy performance.
While these methods demonstrate promising results, they often struggle to generalize across dynamic user behaviors and high-dimensional state-action spaces.

\subsection{DRL for Networks}
DRL has been widely applied to network resource allocation tasks, where diverse user demands and time-varying conditions pose significant challenges to conventional optimization techniques~\cite{shi2023machine}. 
In~\cite{xu2023dofms}, the authors proposed a DRL-based Out-of-Order (OFO)-Friendly Multipath Scheduling (DOFMS) framework to ensure high-bandwidth and low-OFO transmission in mobile HetNets. The framework introduced a novel OFO metric for accurate evaluation, utilized a Double DDQN for dynamic scheduling, and incorporated an asynchronous learning module to handle variable path delays. In~\cite{zhang2023intelligent}, a power control problem was formulated for spectrum-sharing HetNets to improve Global Energy Efficiency (GEE). The authors proposed a DRL-based cloud-edge collaborative framework that allows base stations to independently adjust transmit power, avoiding the scalability limitations of centralized optimization. In~\cite{wang2025tf}, the authors proposed a transformer-enhanced distributed DRL scheduling method for heterogeneous IoT applications across edge and cloud computing environments. The method followed an actor-critic architecture, scaled across multiple distributed servers, and incorporated an off-policy correction mechanism to improve training stability.
Despite these advances, existing DRL approaches often suffer from poor generalization and slow convergence in environments with heterogeneous user behaviors and high-dimensional state spaces.

\subsection{Transformer for Wireless Networks} 
Transformer-based models have emerged as a promising solution for complex wireless networks. In~\cite{zhang2025decision}, the authors proposed a Decision Transformer (DT)-based architecture for wireless resource management, aiming to improve sample efficiency and generalization over conventional DRL. By pre-training in the cloud and fine-tuning at the edge, the DT framework achieved faster convergence and better performance in intelligent reflecting surface-aided and unmanned aerial vehicle-assisted MEC scenarios.
In~\cite{wang2024wireless}, the authors studied deep learning-based wireless interference recognition under non-cooperative electromagnetic attacks. They proposed a multi-modal feature extraction framework with a dual transformer module and adaptive gradient modulation to improve recognition accuracy, convergence speed, and computational efficiency. In~\cite{zhou2024transformer}, a transformer-based model was introduced for Channel State Information (CSI) prediction in high-mobility scenarios. Using multi-head attention and positional encoding, the model effectively captured global channel features and outperformed traditional deep learning models in accuracy with reasonable complexity.
Transformers have also been used to approximate value functions in DRL, including Q-networks and critic networks. For example, the authors in~\cite{esslinger2022deep}  proposed the Deep Transformer Q-Network (DTQN), which employs transformer decoders to encode agent observation histories, yielding more stable and efficient Q-value estimation in partially observable environments. Similarly, the authors in~\cite{tian2025chunking} introduced a transformer-based Soft Actor-Critic (SAC) variant that uses transformers to process $N$-step returns for improved value estimation in complex tasks. The Transformer Off-Policy Episodic RL (TOP-ERL) method~\cite{litop} similarly leverages transformers to estimate value functions from segmented action sequences, improving the handling of long-term dependencies.
Other related works explore planning-inspired architectures. Value Iteration Networks (VIN)~\cite{tamar2016value} embed differentiable planning procedures into neural networks, while the Predictron architecture~\cite{silver2017predictron} conducts internal rollouts to estimate multi-step returns. TreeQN and ATreeC~\cite{farquhar2018treeqn} incorporate tree search mechanisms into value networks to enable forward-looking reasoning. 
While these approaches focus primarily on temporal modeling or multi-step return estimation, {\textit{ReaCritic}} takes a different direction. It introduces a transformer-based critic specifically designed to enhance bidirectional reasoning within single-step value estimation. This design enables more expressive and accurate critic outputs in complex, dynamic environments without relying on full rollout-based return prediction.

\section{System Model}\label{system}
In this section, we present a HetNet system model with heterogeneous mobile users and formulate a multi-resource allocation problem, where users exhibit diverse preferences across different resource optimization objectives.
\begin{figure}[t]
\centering
\includegraphics[width = 0.48\textwidth]{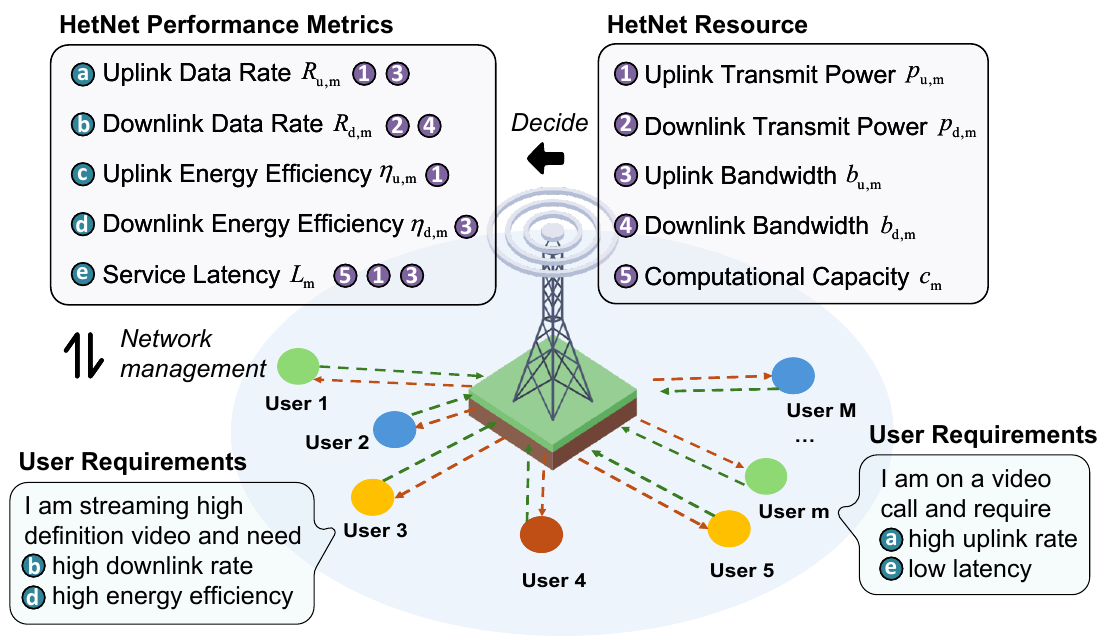}
\caption{System model of the proposed HetNet with one BS and $M$ heterogeneous users characterized by distinct communication demands, computational capabilities, and service
requirement.}
\label{fig_sys}
\end{figure}
\subsection{Wireless Environment}
We consider a HetNet with $M$ heterogeneous mobile users, where each user represents either a real end user or a task-originating agent with specific communication demands, computational capabilities, and service requirements. As illustrated in Fig.~\ref{fig_sys}, the network operates under dynamic and time-varying wireless conditions. The objective is to optimize joint resource allocation strategies, including transmission power, bandwidth, and computation resources, to adaptively satisfy user-specific QoS requirements.

To capture user mobility in a time-varying wireless environment under the coverage of the Base Station (BS), we adopt a random mobility model, where each user's normalized position is updated at every time step via Gaussian perturbation~\cite{yang2020bayesian}: 
\begin{equation}
\begin{aligned}    
&x_m^{(t+1)} = x_m^{(t)} + \Delta x_m^{(t)}, \quad \Delta_{x_m}^{\left(t\right)} \sim \mathcal{N}\left(0, \sigma^2\right),
\end{aligned}
\end{equation}
where $m \in \{1,\ldots,M \}$ and $x_m^{\left(t\right)}$ is the normalized position of the $m_{\rm th}$ user at $t_{\rm th}$ step. This baseline mobility setting captures stochastic user movements and has been widely used for algorithm evaluation in wireless resource management studies~\cite{santi2012mobility}. The parameter $\sigma$ controls the mobility intensity, allowing small random variations within the spatial domain. Each user’s transmission distance $D_m$ is derived from its normalized position and used to compute the large-scale path loss, modeled as
\begin{equation}
\mathrm{\zeta}_m = D_m^{-\gamma},
\end{equation}
where $\gamma$ is the path loss exponent. For small-scale fading, we consider a Rician channel model, which accounts for both line-of-sight and multipath effects. Combining large-scale and small-scale fading, the uplink and downlink SINRs for the $m_{\rm th}$ user, i.e., $\mathrm{SINR}_{{\rm{u}},m}$ and $\mathrm{SINR}_{{\rm{d}},m}$ are defined as
\begin{equation}
\mathrm{SINR}_{{\rm{u}},m} = \frac{p_{{\rm{u}},m} P_{{\rm{u}},{\rm{m}}} \cdot q_{{\rm{u}},m} \cdot \mathrm{\zeta}_m}{\sum_{j \neq m} p_{{\rm{u}},j} P_{{\rm{u}},\max} \cdot q_{{\rm{u}},j} \cdot \mathrm{\zeta}_j + \sigma_{\rm{u}}^2}, 
\end{equation}
and
\begin{equation}
\mathrm{SINR}_{{\rm{d}},m} = \frac{p_{{\rm{d}},m} P_{{\rm{d}},\max} \cdot q_{{\rm{d}},m} \cdot \mathrm{\zeta}_m}{\sum_{j \neq m} p_{{\rm{d}},j} P_{{\rm{d}},\max} \cdot q_{{\rm{d}},j} \cdot \mathrm{\zeta}_j + \sigma_{\rm{d}}^2},
\end{equation}
where $p_{\mathrm{u},m} \in [0, 1]$ and $p_{\mathrm{d},m} \in [0, 1]$ denote the normalized power allocation factors for uplink and downlink transmission. $P_{\mathrm{u},\max}$ is the maximum uplink transmit power of the $m_{\rm th}$ user, while $P_{\mathrm{d},\max}$ is the maximum downlink transmit power determined by the BS. The terms $q_{\mathrm{u},m}$ and $q_{\mathrm{d},m}$ represent the Rician small-scale fading gains for uplink and downlink, and the corresponding noise powers are denoted by $\sigma_{\mathrm{u}}^2$ and $\sigma_{\mathrm{d}}^2$.

Based on the SINR expressions, we derive three key performance metrics for each user in the HetNet: uplink and downlink data rates, energy efficiency, and service latency.

\subsubsection{Data Rate}
The achievable uplink and downlink data rates, i.e., $\mathcal{R}_{{\rm{u}},m}$ and $\mathcal{R}_{{\rm{d}},m}$, are computed using Shannon’s capacity formula:
\begin{equation}
\mathcal{R}_{{\rm{u}},m} = b_{{\rm{u}},m} B_{{\rm{u}},{\rm{m}}} \log_2\left(1 + \mathrm{SINR}_{{\rm{u}},m}\right), 
\end{equation}
and
\begin{equation}
\mathcal{R}_{{\rm{d}},m} = b_{{\rm{d}},m} B_{{\rm{d}},{\rm{m}}} \log_2\left(1 + \mathrm{SINR}_{{\rm{d}},m}\right),
\end{equation}
where $b_{{\rm{u}},m}, b_{{\rm{d}},m} \in [0,1]$ are the normalized bandwidth allocation factors, and $B_{\mathrm{u},m}$ and $B_{\mathrm{d},m}$ are the maximum bandwidths for uplink and downlink transmissions.

\subsubsection{Energy Efficiency}
The uplink and downlink energy efficiencies, denoted by $\eta_{\mathrm{u},m}$ and $\eta_{\mathrm{d},m}$ (in bits per Joule), are defined as the ratio of achievable data rate to transmission power:
\begin{equation}
\eta_{\mathrm{u},m} = \frac{\mathcal{R}_{\mathrm{u},m}}{p_{{\rm{u}},m} P_{{\rm{u}},\max}},
\end{equation}
and
\begin{equation}
\eta_{\mathrm{d},m} = \frac{\mathcal{R}_{\mathrm{d},m}}{p_{{\rm{d}},m} P_{{\rm{d}},\max}},
\end{equation}
where the denominator represents the transmit power allocated to the $m_{\rm th}$ user for uplink and downlink, respectively.

\subsubsection{Service Latency}
We define service latency $L_m$ as the total time required to transmit and process a task with computational demand $d_m$ (in CPU cycles). The first term captures transmission delay, while the second reflects task processing delay as
\begin{equation}
L_m = \frac{{d}_m}{\mathcal{R}_{{\rm{u}},m}} + \frac{{d}_m}{c_m C_{\max} \cdot \rho_m},
\end{equation}
where $c_m \in [0,1]$ is the computational resource allocation ratio for the $m_{\rm th}$ user, $C_{\max}$ is the total computational capacity available in the HetNet, and $\rho_m$ denotes the user’s local computation efficiency, defined as
\begin{equation}
\rho_m = \frac{1}{\kappa_m f_m},
\end{equation}
with $\kappa_m$ representing the effective switched capacitance coefficient determined by the device architecture, and $f_m$ denoting the local CPU frequency of the $m_{\rm th}$ user.

\subsection{Problem Formulation}\label{problem}
The considered multi-resource allocation problem focuses on the HetNet scenarios where heterogeneous users have different preferences over different resource optimization objectives. We use different weight parameters to denote the different preferences. The $m_{\rm th}$ user has specific optimization preferences captured via a weighted utility function $Q_m$, reflecting priorities among uplink and downlink throughput, energy efficiency, and latency. The optimization objective is given as
\begin{equation}\label{eq:opt_problem}
\begin{aligned}
\max_{\left\{ \bm{a}_1, \ldots, \bm{a}_M \right\}} \quad 
& \sum_{m=1}^{M} Q_m\left(\bm{a}_m\right), \\
\text{s.t.} \quad 
& 0 \leq p_{\mathrm{u},m} \leq 1, \\
& 0 \leq p_{\mathrm{d},m} \leq 1, \quad \sum_{m=1}^{M} p_{\mathrm{d},m} \leq 1, \\
& 0 \leq b_{\mathrm{u},m} \leq 1, \quad \sum_{m=1}^{M} b_{\mathrm{u},m} \leq 1, \\
& 0 \leq b_{\mathrm{d},m} \leq 1, \quad \sum_{m=1}^{M} b_{\mathrm{d},m} \leq 1, \\
& 0 \leq c_{m} \leq 1, \quad \sum_{m=1}^{M} c_{m} \leq 1,
\end{aligned}
\end{equation}
where 
\begin{equation}\label{eqactionw}
{\bm{a}}_m = \{p_{{\rm{u}},m}, p_{{\rm{d}},m}, b_{{\rm{u}},m}, b_{{\rm{d}},m}, c_m\}
\end{equation}
denotes the action vector for the $m_{\rm th}$ user, and $Q_m$ is
\begin{align}\label{eq:objective_main_cleaned}
Q_m =\; & \alpha_m^{(1)} \cdot \mathcal{R}_{{\rm{u}},m}\left( p_{{\rm{u}},m},\; b_{{\rm{u}},m} \right) + \alpha_m^{(2)} \cdot \mathcal{R}_{{\rm{d}},m}\left( p_{{\rm{d}},m},\; b_{{\rm{d}},m} \right) \notag \\
& + \alpha_m^{(3)} \cdot \eta_{{\rm{u}},m}\left( p_{{\rm{u}},m}\right) 
+ \alpha_m^{(4)} \cdot \eta_{\rm{d},m}\left( p_{{\rm{d}},m}\right) \notag \\
& - \alpha_m^{(5)} \cdot L_m\left( p_{{\rm{u}},m},\; b_{{\rm{u}},m},\; c_m \right),
\end{align}
where $\alpha_m^{(i)}$, $i \in {1,\ldots,5}$, are user-specific preference weights corresponding to uplink data rate, downlink data rate, uplink energy efficiency, downlink energy efficiency, and service latency, respectively. \textcolor{black}{The specific structure of the power constraints reflects the fundamental difference between uplink and downlink hardware limitations. While downlink transmissions are typically subject to a total power budget at the base station, the uplink power allocation is dictated by the maximum transmit capability of individual user equipment. Therefore, the constraint on $p_{\mathrm{u},m}$ is applied element-wise to characterize the physical power amplifier limits of each mobile terminal. If a total uplink power budget across subcarriers is required in certain implementations, it can be incorporated by adding an additional summation constraint without affecting the proposed ReaCritic architecture.}
The constraints jointly ensure that each user's power, bandwidth, and compute allocations remain within feasible per-user limits, while also satisfying global capacity constraints across all users.

\section{Reasoning Transformer-based DRL Critic-model Scaling}\label{large}
In this section, we formulate the MDP for the DRL-based HetNet management and introduce {\textit{ReaCritic}}, a reasoning transformer-based critic-model scaling scheme designed to address key challenges in HetNets.

\subsection{DRL Markov Decision Process Formulation}
DRL offers a principled framework for sequential decision-making by enabling an agent to interact with an environment $\mathcal{E}$. 
At each discrete time step $t $, the agent observes a state $ {\bm{s}}_t \in \mathcal{S} $ from the state space and selects an action $ {\bm{a}}_t \in \mathcal{A} $ according to its learned policy. The environment responds with a reward signal $ r_t^{\mathcal{E}} \in \mathbb{R} $, and transitions to a new state $ {\bm{s}}_{t+1} $ based on a transition function $ T: \mathcal{S} \times \mathcal{A} \rightarrow \mathcal{S} $. The learning objective is to optimize the expected cumulative reward by modeling the agent–environment interaction as an MDP.

\begin{itemize}
\item \textbf{Environment:} 
The environment $\mathcal{E}$ captures the dynamics of HetNets, including user mobility, time-varying wireless channels, interference, and fluctuations in communication and computation resources. 
It enforces global constraints on power, bandwidth, and computational capacity. 
At each time step $t$, the DRL agent receives user-specific observations such as channel quality, task demand, latency requirement, position, computation capability, and user type. The environment computes uplink and downlink rates, evaluates task latency, and provides a multi-objective reward signal that reflects a weighted combination of throughput, energy efficiency, and service latency.

\item \textbf{State:} 
The state of the $m_{\rm th}$ user at time $t$ is represented by a high-dimensional vector:
\begin{align}
{\bm{s}}_t^m =& \{d_m, L_m, x_m, f_m, \tau_m, P_{{\rm{u}},{\rm{m}}}, q_{{\rm{u}},m}, \mathrm{\zeta}_m, \sigma_{\rm{u}}, \notag \\
&
P_{{\rm{d}},{\max}}, q_{{\rm{d}},m}, \sigma_{\rm{d}}, C_{\max}, \rho_m \},
\end{align}
where $\tau_m \in \{0, 1, \ldots, \tau_{\max}\}$ denotes the user type.Each user type corresponds to a distinct set of preferences over multiple objectives, encoded by weights $\alpha_m^{(i)}$ $\left(i=1,\ldots,5\right)$ in the utility function $Q_m$. These profiles are hidden from the network manager, introducing uncertainty into the learning process and increasing the complexity of policy adaptation.

The global state at time $t$ is obtained by concatenating all user states as
\begin{equation}
{\bm{s}}_t = \left\{ {\bm{s}}_t^1, {\bm{s}}_t^2, \ldots, {\bm{s}}_t^M \right\}
\end{equation}
forming a joint high-dimensional input space that scales linearly with the number of users $M$.

This high-dimensional representation captures a rich set of features that reflect both individual-level heterogeneity and global system dynamics, which are essential for making fine-grained, context-aware resource allocation and scheduling decisions in HetNets. However, such complexity also introduces significant challenges for DRL-based methods, including increased sample complexity, potential overfitting, and difficulties in generalization. 
Therefore, advanced policy and critic architectures are necessary to effectively model and extract useful patterns from this high-dimensional, structured state space.

\item \textbf{Action}
The action vector for the $m_{\rm th}$ user is defined based on the key optimization variables in the objective formulation as \eqref{eqactionw}.

At each time step $t$, the DRL agent jointly selects the global action vector as
\begin{equation} 
\bm{a}_t = \left\{ \bm{a}_t^1, \bm{a}_t^2, \dots, \bm{a}_t^M \right\}.
\end{equation} 

\item \textbf{Reward}
The final reward for each user is composed of a weighted sum:
\begin{equation}
\begin{aligned}
r_m &= \alpha_m^{(1)}\mathcal{R}_{u,m} + \alpha_m^{(2)} \mathcal{R}_{d,m}
- \alpha_m^{(3)} b_{u,m}
- \alpha_m^{(4)} b_{d,m}\\
&- \alpha_m^{(5)} p_{u,m}^2
- \alpha_m^{(6)} p_{d,m}^2
- \alpha_m^{(7)} \cdot \mathbb{I}\left(l_m > l_{\mathrm{th},m}\right),
\end{aligned}
\end{equation}
where $\alpha_m^{(i)}$, $i \in \{1,2,3,4,5\}$ are task-specific weight coefficients. $l_m$ represents the total latency experienced by the $m_{\rm th}$ user, and $\mathbb{I}\left(l_m > l_{\mathrm{th},m}\right)$ is an indicator function penalizing violations of latency constraints. $l_{\mathrm{th},m}$ is the latency requirement for the $m_{\rm th}$ user.
\end{itemize}
This MDP formulation enables the DRL agent to learn preference-aware scheduling strategies in highly complex HetNet environments with heterogeneous users. The joint state and action spaces are high-dimensional due to per-user variability and system-wide coupling. For instance, with $M = 50$ users and each user state vector comprising $14$ elements, e.g., task demand, position, fading, noise, and compute capacity, the global state dimension reaches $700$. Similarly, with five decision variables per user, the joint action space becomes $250$-dimensional. This scale, combined with non-stationary dynamics, latent user preferences, and inter-user interference, poses substantial challenges for policy learning, generalization, and sample efficiency.

\subsection{{\textit{ReaCritic}} Design}
\begin{figure}[t]
\centering
\includegraphics[width = 0.45\textwidth]{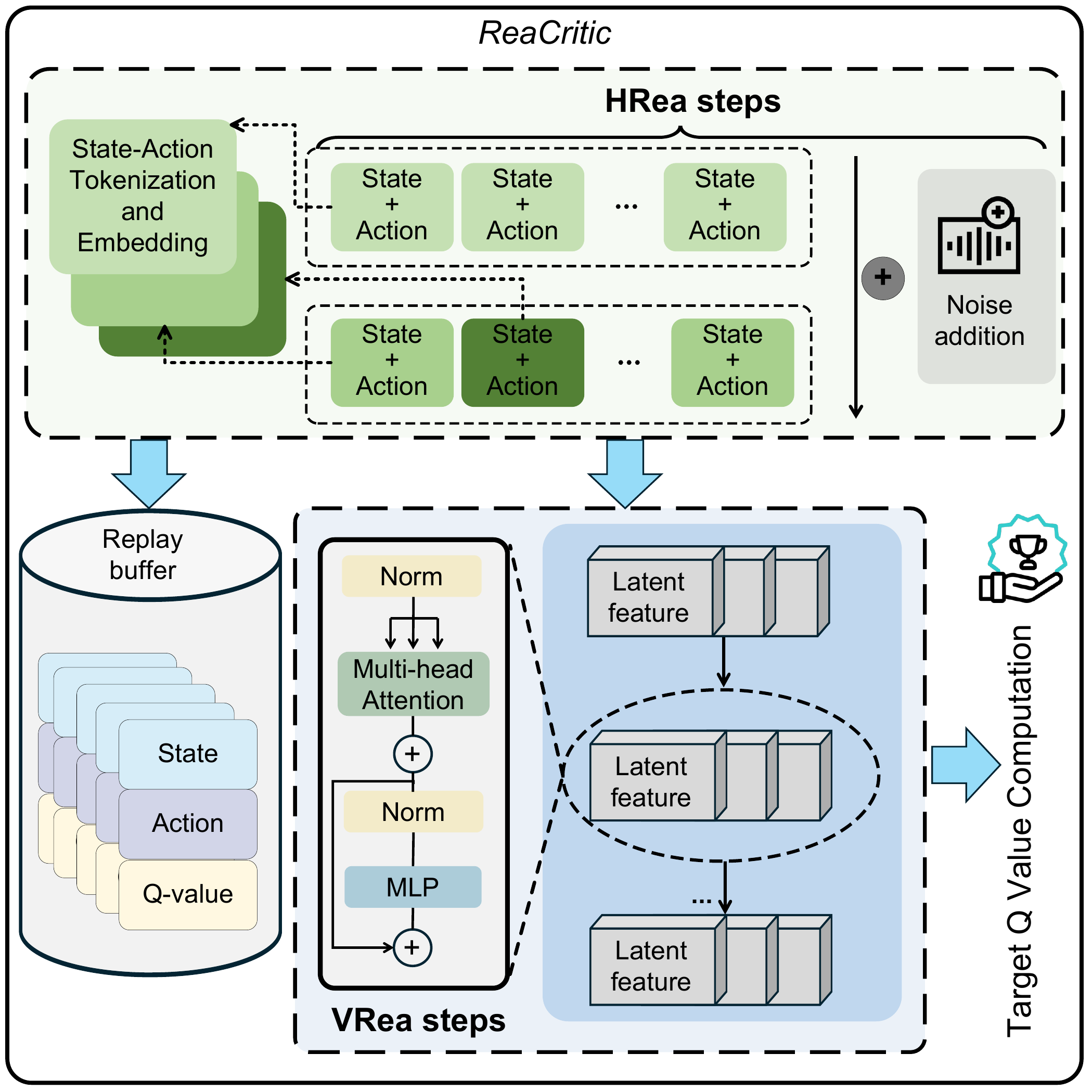}
\caption{The architecture of the proposed \textit{ReaCritic}, which integrates two-dimensional reasoning: horizontal expansion via HRea steps to enrich token diversity, and vertical abstraction via VRea steps to enhance hierarchical value representation.}
\label{fig_tramsformer}
\end{figure}
\textit{ReaCritic} introduces a two-dimensional reasoning architecture designed to enhance the expressiveness and generalization capability of the critic network. \textcolor{black}{Unlike conventional single-stream Transformer critics, our architecture is characterized by a structural modification that instantiates parallel reasoning paths. This design allows the framework to transcend static function approximation, which is otherwise limited to the (1,1) configuration where $H=1$ and $V=1$ represents a standard Transformer-based baseline.} The first stage, denoted as \textbf{Horizontal Reasoning} (HRea), performs token-level expansion by generating multiple parallel state-action embeddings to simulate diverse perspectives. \textcolor{black}{This mechanism is structurally distinct from standard ensemble methods; rather than training multiple independent }{\color{black} networks, HRea expands the representational breadth within a single forward pass by constructing parallel attention branches that process replicated and diversified tokens simultaneously.} The second stage, denoted as \textbf{Vertical Reasoning} (VRea), applies stacked Transformer blocks to process these tokens hierarchically. \textcolor{black}{This vertical stacking facilitates hierarchical logic distillation across the parallel branches generated by HRea, enabling the model to refine complex dependencies through recursive abstraction rather than simple layer stacking.}

As illustrated in Fig. 3, the HRea and VRea steps together enable \textit{ReaCritic} to support fine-grained relational modeling and robust Q-value estimation in complex, high-dimensional HetNets. The whole procedure is provided in Algorithm 1. \textcolor{black}{The architectural blueprint of ReaCritic is predicated on the Transformer mechanism to encapsulate the complex global dependencies inherent in highly coupled HetNets. Unlike convolutional or recurrent structures, which are often constrained by localized receptive fields or sequential processing bottlenecks, the self-attention mechanism facilitates a permutation-invariant modeling of all-to-all correlations among heterogeneous users. This structural property is essential for representing the intricate interference regimes where the operational state of a single mobile terminal significantly influences the global network utility. Consequently, the adoption of a transformer-based backbone is a principled choice to ensure that the Critic network possesses the necessary structural capacity to approximate the high-dimensional value functions of modern wireless environments.}

\textcolor{black}{Furthermore, the synergy between horizontal and vertical reasoning represents a systematic orchestration of inference-time compute rather than an empirical aggregation of disparate modules. The HRea module is designed to expand the representational breadth through parallel tokens to explore diverse state-action correlations, whereas the VRea module enables the hierarchical distillation of logical abstractions across stacked layers. This dual-axis scaling methodology ensures that the computational effort of the Critic is adaptively aligned with the intrinsic complexity of the task, thereby providing a robust and theoretically grounded pathway for achieving value estimation stability. By shifting the focus from fixed-capacity modeling to dynamic reasoning-time scaling, the proposed architecture transcends the limitations of static empirical stacking and offers a scalable solution for resource management in non-stationary 5G/6G scenarios.}
\begin{algorithm}[t]
\caption{{\color{black}\textit{ReaCritic}-based DRL Framework}}
\label{alg:ReaCritic}
\hspace*{0.02in} {\bf Input}
\begin{itemize}
\item[\textbullet] \textit{Environment} Markov environment $\mathcal{E}$, replay buffer $\mathcal{D}$
\item[\textbullet] \textit{DRL Algorithm $\mathcal{A}$} discount factor $\gamma$, actor network $\pi_\theta$ (if have), target networks $Q_{\bar{\phi}}$, $\pi_{\bar{\theta}}$, learning rates $\eta_Q, \eta_\pi$, mini-batch size $M$
\item[\textbullet] \textit{ReaCritic} critic network $Q_\phi$ with parameters $\phi$, HRea steps $H$, VRea steps $V$
\end{itemize}
\begin{algorithmic}[1]
\State Initialize policy network $\pi_\theta$, critic network $Q_\phi$, and replay buffer $\mathcal{D}$
\For{each training iteration}
\State \textit{ Collect experience}
\State Reset environment and observe initial state $s_0$
\While{not terminal}
\State Sample action $a_t \sim \pi_\theta(\cdot|s_t)$ according to DRL algorithm $\cal A$ policy rules
\State Execute $a_t$, observe $r_t$, $s_{t+1}$, and termination flag $d_t \in {0, 1}$
\State Store transition $(s_t, a_t, r_t, s_{t+1}, d_t)$ in $\mathcal{D}$
\State Update $s_t \leftarrow s_{t+1}$, $t \leftarrow t + 1$
\EndWhile

\State \textit{ Update phase}
\State Sample mini-batch ${(s_t, a_t, r_t, s_{t+1})}$ from $\mathcal{D}$

\State \textcolor{black}{\textbf{(Module 1. Horizontal Reasoning Module)}}
\State \textcolor{black}{Compute state-action embedding $\mathbf{z}_0$ using~\eqref{faefe}}
\State \textcolor{black}{Instantiate parallel paths by expanding into $H$ tokens using~\eqref{zofe} and add exploration noise via~\eqref{felais}}

\State \textcolor{black}{\textbf{(Module 2. Vertical Reasoning Module)}}
\State \textcolor{black}{Perform hierarchical logic distillation through $V$ transformer blocks using~\eqref{vreastepss}}
\State \textcolor{black}{Synthesize representations via step-wise attention~\eqref{combinesf} and predict final Q-value using~\eqref{finalQs}}

\State \textbf{(Critic Loss and Parameter Update)}
\State Compute target Q-value $y_t$
\begin{equation*}
y_t =
\begin{cases}
r_t + \gamma Q_{\bar{\phi}}(s_{t+1}, \pi_{\bar{\theta}}(s_{t+1})), & \text{actor-critic} \\
r_t + \gamma \max_{a'} Q_{\bar{\phi}}(s_{t+1}, a'), & \text{Q-learning}
\end{cases}
\end{equation*}
\State Compute critic loss $\mathcal{L}Q = \frac{1}{M} \sum{i=1}^M (Q_\phi(s_t, a_t) - y_t)^2$
\State Update critic parameters $\phi \leftarrow \phi - \eta_Q \nabla_\phi \mathcal{L}_Q$

\If{actor $\pi_\theta$ exists}
\State \textcolor{black}{Compute policy gradient based on the reasoned Q-estimates using~\eqref{policy}}
\State Update actor \quad $\theta \leftarrow \theta + \eta_\pi \nabla_\theta J$
\EndIf
\EndFor
\end{algorithmic}
\hspace*{0.02in} \textbf{Output} Trained DRL algorithm $\cal A$
\end{algorithm}

\subsubsection{Horizontal Reasoning Path}
The core motivation behind the HRea design originates from the success of inference-time scaling in LLMs. In LLMs, performance improves when more intermediate tokens are generated during reasoning, e.g., Chain-of-Thought (CoT), enabling the model to engage in multi-step deduction before reaching a final output. While the critic in DRL is not tasked with next-token prediction, it performs complex mappings from high-dimensional state-action pairs to value estimates. These mappings can benefit from intermediate representations that act as latent reasoning anchors, similar in spirit to the token sequences in LLMs.

However, unlike LLMs that sequentially generate tokens, DRL critic networks operate under a single-shot evaluation paradigm. 
Therefore, to mimic multi-step reasoning, \textit{ReaCritic} \textcolor{black}{employs a horizontal reasoning expansion strategy whereby the model replicates the input state-action token multiple times} and augments each copy with a unique positional embedding, effectively simulating parallel reasoning paths.

Let the input state and action pair be $\bm{x} \in \mathbb{R}^{B \times d_{\bm{s}}}$ and $\bm{a} \in \mathbb{R}^{B \times d_{\bm{a}}}$ for batch size $B$. We denote $\bm{x} \| \bm{a}$ as the concatenation of the observation vector $\bm{x} \in \mathbb{R}^{B \times d_{\bm{s}}}$ and the action vector $\bm{a} \in \mathbb{R}^{B \times d_{\bm{a}}}$ along the feature dimension, resulting in a joint representation of size $\mathbb{R}^{B \times (d_{\bm{s}} + d_{\bm{a}})}$. This concatenated input captures both environment state and decision action, and serves as the input to the critic's embedding module. We define the base embedding through a linear projection followed by normalization as
\begin{equation}\label{faefe}
\mathbf{z}_0 = \mu\left(\mathbf{W}_{\text{embed}} \left[ \bm{x} \| \bm{a} \right]\right) \in \mathbb{R}^{B \times d_{\rm h}},
\end{equation}
where $\mathbf{W}{\text{embed}} \in \mathbb{R}^{(d{\bm{s}} + d_{\bm{a}}) \times d_{\rm h}}$ is the trainable projection matrix, and $\mu(\cdot)$ denotes layer normalization. To enable multi-path reasoning, the base embedding is expanded into a sequence of $H$ horizontally parallel reasoning tokens as
\begin{equation}\label{zofe}
\mathbf{Z}^{(0)} = \left[ \mathbf{z}_0 + \mathbf{e}_1, \ldots, \mathbf{z}_0 + \mathbf{e}_H \right] \in \mathbb{R}^{B \times H \times d_{\rm h}},
\end{equation}
where $\mathbf{e}_i$ is the learnable positional encoding for the $i_{\text{th}}$ HRea step. To promote diversity and encourage exploration during training, an optional Gaussian noise term $\mathcal{N}(0, \sigma^2)$ can be added to each token:
\begin{equation}\label{felais}
\mathbf{Z}^{(0)} \leftarrow \mathbf{Z}^{(0)} + \mathcal{N}(0, \sigma^2).
\end{equation}
This horizontally extended token sequence is then passed to the subsequent vertical reasoning transformer layers, where latent value estimation is refined through attention-based abstraction. \textcolor{black}{Intuitively, the expansion mechanism defined in Equations 19 through 21 simulates the generation of parallel thought paths, where each reasoning token explores a unique representational subspace of the state-action pair to mitigate the risks of localized feature bias in high-dimensional HetNet environments.} The key intuition behind HRea is to provide the critic network with multiple perspectives on the same decision point, enhancing robustness, exploration, and its ability to model high-order interactions in complex environments. \textcolor{black}{This mechanism systematically broadens the attention span of the Critic, allowing it to evaluate multiple potential interference patterns simultaneously rather than relying on a collapsed mean-field approximation typically found in standard architectures.}

\subsubsection{Vertical Reasoning Path}
While HRea expands a single embedding point into multiple hypothetical reasoning paths, each token in this set still represents a shallow projection of the original state-action pair. To perform deeper reasoning and generate higher-order decision insights, we introduce VRea as a hierarchical processing stack.

This design is motivated by the idea of latent multi-step reasoning in LLMs~\cite{geiping2025scaling}, where deep transformer layers repeatedly refine representations in hidden space to support complex, context-dependent computation. Here, VRea does not explicitly produce symbolic CoT but instead performs recursive abstraction within the vector space. This allows the critic to simulate structured inference trajectories by reprocessing the latent input across multiple transformer layers, enabling deeper exploration and more robust Q-value modeling. \textcolor{black}{Unlike simple feed-forward stacking found in standard models, this recursive distillation process ensures that the architectural depth translates directly into logical expressiveness across the parallel branches, effectively bridging the gap between high-level reasoning and low-level value estimation.} Additionally, VRea also shares conceptual links with diffusion policy~\cite{chi2023diffusion} and iterative policy refinement, where each step in the stack can be interpreted as an implicit planning operation conditioned on both prior layers and current observations. Compared with simple feed-forward architectures, the VRea stack increases reasoning depth without increasing parameter count proportionally, and facilitates richer representation learning through hierarchical feature composition.

Concretely, the output of HRea, denoted as $\mathbf{Z}^{(0)} \in \mathbb{R}^{B \times H \times d_{\rm h}}$, is processed by a stack of $V$ transformer blocks. Each transformer block $\Psi^{(v)}$ consists of a multi-head self-attention module followed by a feed-forward MLP layer, both wrapped with layer normalization and residual connections in a pre-norm configuration. This stacked vertical structure is implemented as
\begin{equation}\label{vreastepss}
\mathbf{Z}^{(v+1)} = \Psi^{(v)}\left(\mathbf{Z}^{(v)}\right), \quad v = 0, \ldots, V-1,
\end{equation}
where $\Psi^{(v)}(\cdot)$ denotes the $v_{\rm th}$ transformer block. Each layer refines the horizontal token sequence by integrating long-range dependencies and capturing high-order interactions through attention and non-linear transformation. \textcolor{black}{Specifically, while HRea focuses on representational breadth, the recursive operations in Equations 22 and 23 facilitate hierarchical logic distillation by iteratively filtering redundant information and accentuating critical dependencies necessary for precise value estimation.} Additionally, after $V$ layers, the final representation $\mathbf{Z}^{(V)} \in \mathbb{R}^{B \times H \times d_{\rm h}}$ is aggregated using attention over the $H$ horizontal reasoning steps as
\begin{equation}\label{combinesf}
\hat{\mathbf{z}} = \sum_{i=1}^{H} \frac{
\exp\left( \left(\mathbf{Z}^{(V)} \cdot \mathbf{w}_a\right)_i \right)
}{
\sum_{j=1}^{H} \exp\left( \left(\mathbf{Z}^{(V)} \cdot \mathbf{w}_a\right)_j \right)
} \cdot \mathbf{Z}^{(V)}\left[:, i, :\right],
\end{equation}
where $\mathbf{w}_a$ is a learned projection. The final Q-value estimate is obtained as
\begin{equation}\label{finalQs}
Q(\bm{x}, \bm{a}) = \mathbf{w}_q^\top \cdot \mu(\hat{\mathbf{z}}),
\end{equation}
where $\mathbf{w}_q \in \mathbb{R}^{d_{\rm h}}$ is a trainable projection vector used to map the final latent representation to a scalar Q-value.

This vertical reasoning mechanism enables \textit{ReaCritic} to synthesize structured value estimates by abstracting state-action pairs, resulting in improved generalization and temporal stability across complex HetNet scenarios. \textcolor{black}{To further ensure technical clarity, we have introduced explicit modular labels within Algorithm 1 to distinguish between the Horizontal and Vertical reasoning stages. Each module is now preceded by structural comments that delineate the input-output flow, providing a transparent roadmap for implementing the proposed inference-time scaling strategy in large-scale wireless resource management tasks.}

\subsubsection{Integration with DRL algorithms}
\begin{figure*}[t]
\centering
\includegraphics[width=0.9\textwidth]{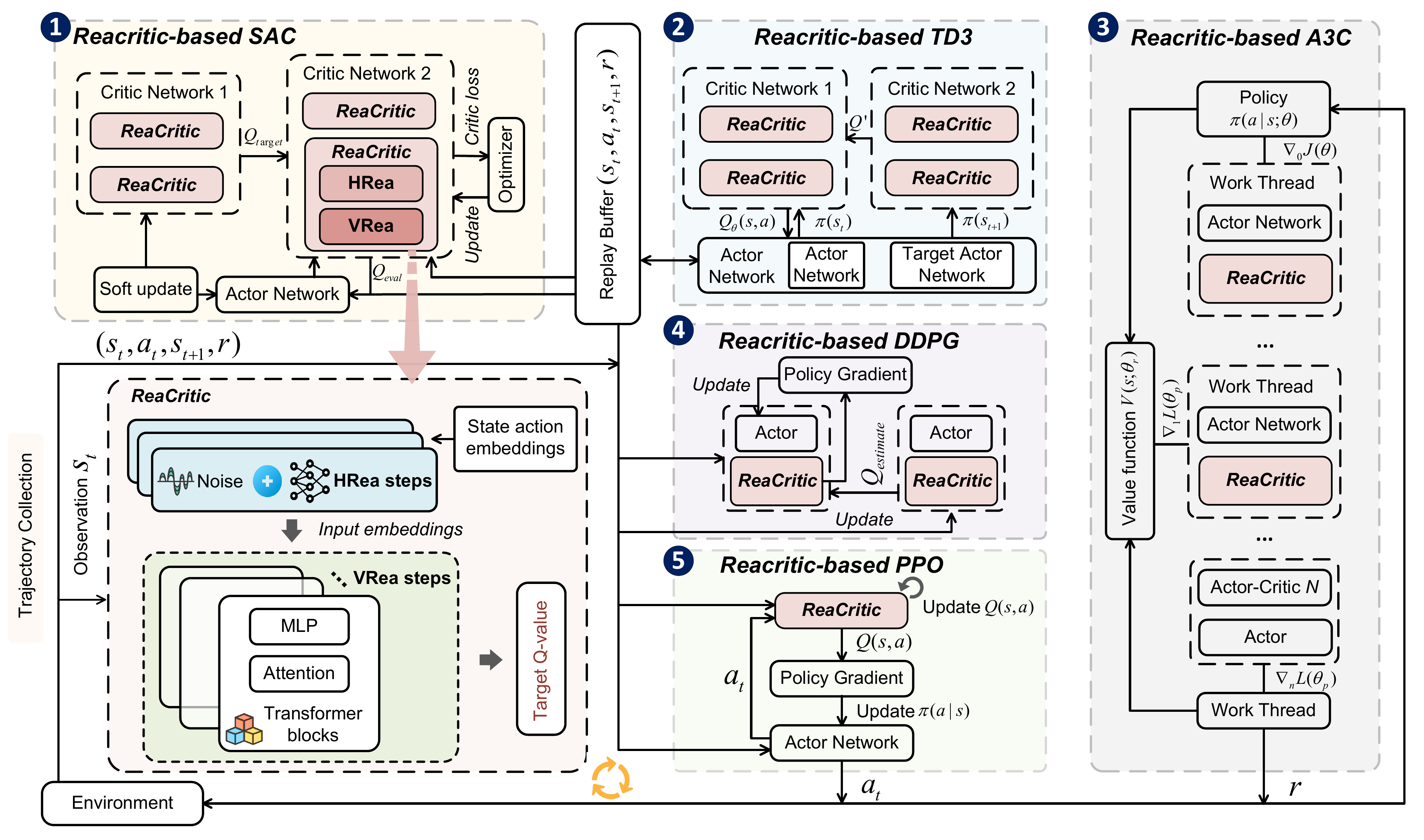}
\caption{Integration of \textit{ReaCritic} into DRL algorithms. Part 1 illustrates the reasoning-augmented SAC pipeline. Parts 2 to 5 demonstrate compatibility with other actor–critic methods, including TD3, A3C, DDPG, and PPO.}
\label{fig_drlrea}
\end{figure*}
As shown in Fig.~\ref{fig_drlrea}, \textit{ReaCritic} is designed as a plug-and-play critic module compatible with standard actor-critic DRL algorithms. While our implementation is based on the SAC framework~\cite{haarnoja2018soft}, the design can be generalized to other value-based methods such as Twin Delayed Deep Deterministic (TD3)~\cite{fujimoto2018addressing}, Deep Deterministic Policy Gradient (DDPG)~\cite{lillicrap2015continuous}, and Asynchronous Advantage Actor-Critic (A3C) algorithms~\cite{babaeizadeh2016reinforcement}.

At each interaction step with the environment, the agent observes a state $\bm{s}_t$, samples an action $\bm{a}_t \sim \pi_\theta\left(\bm{s}_t\right)$, receives a scalar reward $r_t$, and transitions to a new state $\bm{s}_{t+1}$. The tuple $(\bm{s}_t, \bm{a}_t, r_t, \bm{s}_{t+1})$ is stored in the replay buffer $\mathcal{D}$.

To estimate Q-values, \textit{ReaCritic} first constructs a joint state-action embedding using~\eqref{faefe}, which is then expanded into $H$ horizontally reasoned tokens via~\eqref{zofe} and~\eqref{felais}. This horizontal sequence $\mathbf{Z}^{(0)}$ is refined through $V$ vertically stacked transformer blocks as described in~\eqref{vreastepss}. After vertical reasoning, the refined tokens are aggregated across reasoning steps using attention as in~\eqref{combinesf}. The final Q-value is then produced by a lightweight output head as defined in~\eqref{finalQs}.

The critic is trained by minimizing the Bellman residual~\cite{piot2014boosted} as
\begin{equation}
\mathcal{L}_Q = \mathbb{E}_{(\bm{s}, \bm{a}, r, \bm{s}')} \left[ \left(Q_\phi(\bm{s}, \bm{a}) \!-\! \left(r \!+\! \gamma Q_{\phi'}(\bm{s}', \pi_\theta(\bm{s}')) \right)\right)^2 \right].
\end{equation}

The actor is updated to maximize the expected Q-value:
\begin{equation} \label{policy}
\nabla_\theta J(\pi) = \mathbb{E}_{\bm{s} \sim \mathcal{D}} \left[ \nabla_{\bm{a}} Q_\phi(\bm{s}, \bm{a})\big|_{\bm{a} = \pi_\theta(\bm{s})} \cdot \nabla_\theta \pi_\theta(\bm{s}) \right].
\end{equation}
By embedding inference-time reasoning into the critic, \textit{ReaCritic} enhances learning stability, improves Q-value estimation accuracy, and accelerates policy convergence in complex wireless environments. The whole process is shown in Algorithm~\ref{alg:ReaCritic}

\subsubsection{Theoretical Complexity and Convergence Analysis}
Introducing \textit{ReaCritic} increases the expressive power of the critic network at the cost of additional computational complexity. Compared to standard feedforward critic models, which compute a single forward pass through an MLP, \textit{ReaCritic} employs a Transformer-based dual-axis reasoning architecture with $H$ horizontal tokens and $V$ stacked reasoning layers.

The per-update computational complexity of \textit{ReaCritic} is $\mathcal{O}(B H^2 d_{\rm h} + B H V d_{\rm h}^2)$, where $B$ is the batch size, $H$ is the number of HRea steps, $V$ is the number of vertical Transformer blocks, and $d_{\rm h}$ is the hidden dimension. The first term corresponds to the cost of multi-head self-attention across $H$ tokens per layer, while the second term captures the complexity of feed-forward transformations in each Transformer block. This structure scales linearly with $B$ and $V$, and quadratically with $H$ and $d_{\rm h}$. Compared to standard MLP-based critics with complexity $\mathcal{O}(B d_{\rm h}^2)$, the added cost remains moderate for lightweight settings and can be efficiently batched on modern GPUs. \textcolor{black}{The rationale for this increased complexity resides in the necessity of representational breadth, as large-scale HetNets require the structural capacity to explore vast state-action correlations that traditional MLPs cannot effectively capture. This overhead is mitigated by the architectural synergy with modern AI hardware, where SIMD (Single Instruction, Multiple Data) parallelization ensures that the increase in wall-clock time remains sub-linear relative to theoretical FLOPs. Consequently, ReaCritic favors scalability by processing multiple reasoning tokens simultaneously, justifying the marginal latency increase through significant gains in convergence stability.}

\textcolor{black}{Furthermore, the hyperparameters $H$ and $V$ offer a controllable trade-off between logical depth and representation space. While increasing the vertical depth $V$ scales} {\color{black}computational cost linearly for enhanced abstraction, scaling the horizontal breadth $H$ expands the exploration of state-action patterns through parallel attention heads without a prohibitive penalty on sequential processing time. This structural flexibility allows network operators to dynamically adjust the reasoning-time compute of ReaCritic to match available hardware resources and the non-stationary demands of the HetNet environment, ensuring the framework remains practically tractable for large-scale deployments.}
More importantly, \textit{ReaCritic} is designed to support scaling along both the depth ($V$) and breadth ($H$) dimensions, enabling deployment of reasoning-capable critics. This two-dimensional scaling path aligns with recent trends in inference-time compute augmentation and allows the model to serve both compact on-device learning and large-scale centralized training when compute resources permit.

Regarding convergence, although \textit{ReaCritic} increases model depth and nonlinearity, its use of pre-normalized residual blocks and standard Transformer modules facilitates stable training.
Since it is integrated into conventional DRL update rules, e.g., Bellman backup and policy gradients, the theoretical convergence properties of the base algorithms, such as SAC or TD3, remain valid, assuming the critic maintains universal approximation capacity. The added reasoning depth improves the expressiveness of the value function, which can reduce approximation bias and accelerate convergence. Additionally, stepwise attention and horizontal perturbation act as implicit regularization, improving stability and credit assignment in complex, non-stationary environments.

\textcolor{black}{From a deployment perspective, the feasibility of ReaCritic within 5G/6G ecosystems is governed by the critical trade-off between signaling overhead, coordination stability, and execution latency. First, regarding the communication footprint, the centralized training paradigm is essential to mitigate the non-stationarity inherent in coupled wireless environments. Although aggregating global state information incurs a signaling overhead during the learning phase, this investment is justified by the resulting stability gain, which enables the network to converge toward a robust joint resource allocation policy that suppresses mutual interference. Crucially, this centralization does not preclude real-time responsiveness. The architecture supports a decoupled execution model where the trained Actor is deployed for distributed inference at individual access points. This ensures that the sophisticated reasoning of the Critic guides the optimization process without imposing communication bottlenecks during the ultra-low latency transmission phase.}

\textcolor{black}{Second, regarding computational complexity, the overhead of reasoning-time scaling is systematically manageable through hardware-software synergy. Although the theoretical complexity scales quadratically with the horizontal breadth $H$, the multi-head attention mechanism is inherently amenable to parallelization. By leveraging dedicated edge AI accelerators or gNodeB-integrated tensor cores, the execution time for standard configurations can be maintained within the sub-millisecond thresholds of the 5G Transmission Time Interval (TTI). Furthermore, the structural flexibility of ReaCritic allows network operators to dynamically adjust the reasoning depth $V$ and breadth $H$ to match available hardware resources, ensuring that the framework remains both practically tractable and theoretically sound for large-scale deployments.}


\section{Numerical Results}\label{num}
We evaluate the effectiveness and generalizability of \textit{ReaCritic} in two settings: (i) a custom-designed HetNet environment simulating dynamic wireless conditions and heterogeneous user demands, and (ii) standard DRL control benchmarks from OpenAI Gym to assess cross-domain applicability. All experiments are conducted on a workstation running Ubuntu 20.04, equipped with an AMD EPYC 7543 32-core CPU and 4 NVIDIA RTX 4060 GPUs.

\textcolor{black}{The evaluation framework is meticulously curated to encapsulate the multifaceted technical constraints of next-generation communication systems while maintaining algorithmic transparency. In particular, the custom HetNet environment replicates a dense deployment of macro and small-cell base stations, characterized by high-dimensional coupling and complex interference regimes. To encapsulate the physical layer intricacies of urban 5G/6G deployments, we incorporate Rician fading to model multipath propagation and Gaussian mobility models to account for the stochastic nature of user trajectories. These elements collectively simulate the environmental non-stationarity and unpredictability that challenge conventional learning paradigms, thereby providing a rigorous platform to evaluate the reasoning-time scaling capabilities of ReaCritic under realistic network dynamics.}

\subsection{Experimental Setup}
\subsubsection{Wireless Benchmark Environment}
We construct a high-dimensional wireless environment that simulates a HetNet with $M \in \left[10, 50\right]$ users. As the number of users increases, the decision space expands significantly, making per-user resource allocation highly complex and inefficient. In this multi-user scenario, we consider a network central node that makes global decisions for all users. 
In this setting, each user is randomly located in a normalized two-dimensional area and exhibits independent mobility modeled by Gaussian perturbations at each time step. Channel conditions follow a Rician fading model combined with log-distance path loss, with transmission distances sampled from $\left[5, 10\right]$ meters. Channel gains are updated in each episode to capture temporal variation.

Each user is characterized by heterogeneous attributes, including task demand, latency requirement, computation capability, channel quality, physical location, and a one-hot encoded type indicator. 
At each decision step, the agent selects a five-dimensional action vector per user that determines uplink power, downlink power, bandwidth, and computation allocation. Actions are normalized and clipped to satisfy global resource constraints. The environment then computes uplink and downlink rates, total task latency, and a multi-objective reward that integrates throughput, energy efficiency, resource utilization, and latency satisfaction. 

\subsubsection{DRL Benchmark Environments}
\textcolor{black}{Furthermore, we utilize standardized benchmarks from the OpenAI Gym library to assess the generalizability of the proposed framework and to ensure experimental reproducibility. High-dimensional continuous control tasks such as HumanoidStandup-v4 are selected due to their requirement for intricate coordination across highly dependent state variables, which serves as a functional proxy for the resource allocation complexities inherent in wireless networks. By demonstrating consistent performance gains across both domain-specific wireless scenarios and established control benchmarks, we substantiate the robust architectural versatility of our dual-axis reasoning mechanism. This multifaceted evaluation confirms that the efficacy of horizontal and vertical reasoning is both practically relevant to wireless systems and theoretically generalizable to broader classes of high-dimensional optimization problems across varying levels of environmental complexity.} To evaluate the general applicability of \textit{ReaCritic}, we conduct experiments across two categories of benchmark environments.
First, we compare against standard DRL algorithms, i.e., SAC, DDPG, A3C, TD3, and PPO, within our custom wireless resource allocation environment. These serve as strong baselines for measuring learning stability, sample efficiency, and final policy performance.
Second, to assess task generalizability beyond wireless systems, we integrate \textit{ReaCritic} into these DRL algorithms and test them on three representative OpenAI Gym~\cite{brockman2016openai} control tasks with varying complexity:
\begin{itemize}
\item \textbf{MountainCarContinuous-v0:} A low-complexity continuous control task where the agent must build momentum by swinging back and forth to drive a car up a steep hill.
\item \textbf{HumanoidStandup-v4:} A medium-complexity task where a quadruped robot learns to stand and maintain balance by coordinating multiple joints.
\item \textbf{Ant-v4:} A high-complexity locomotion task involving a multi-jointed humanoid robot that must learn stable standing and walking behaviors through high-dimensional action coordination.
\end{itemize}
These environments provide diverse and challenging domains to validate \textit{ReaCritic}'s ability to enhance value estimation and policy learning across different settings.

\subsection{Experiments Performance Analysis}
\subsubsection{Performance Evaluation in HetNet with Varying User Density}
We evaluate how \textit{ReaCritic}-based SAC performs under different numbers of users in the proposed HetNet environment, and compare it with standard SAC in terms of convergence behavior and Q-value estimates.

\begin{figure*}[t]
\centering
\subfigure[HRea step number is 1, VRea step number is 1.]
{
\includegraphics[width=0.23\textwidth, height=0.18\textwidth]{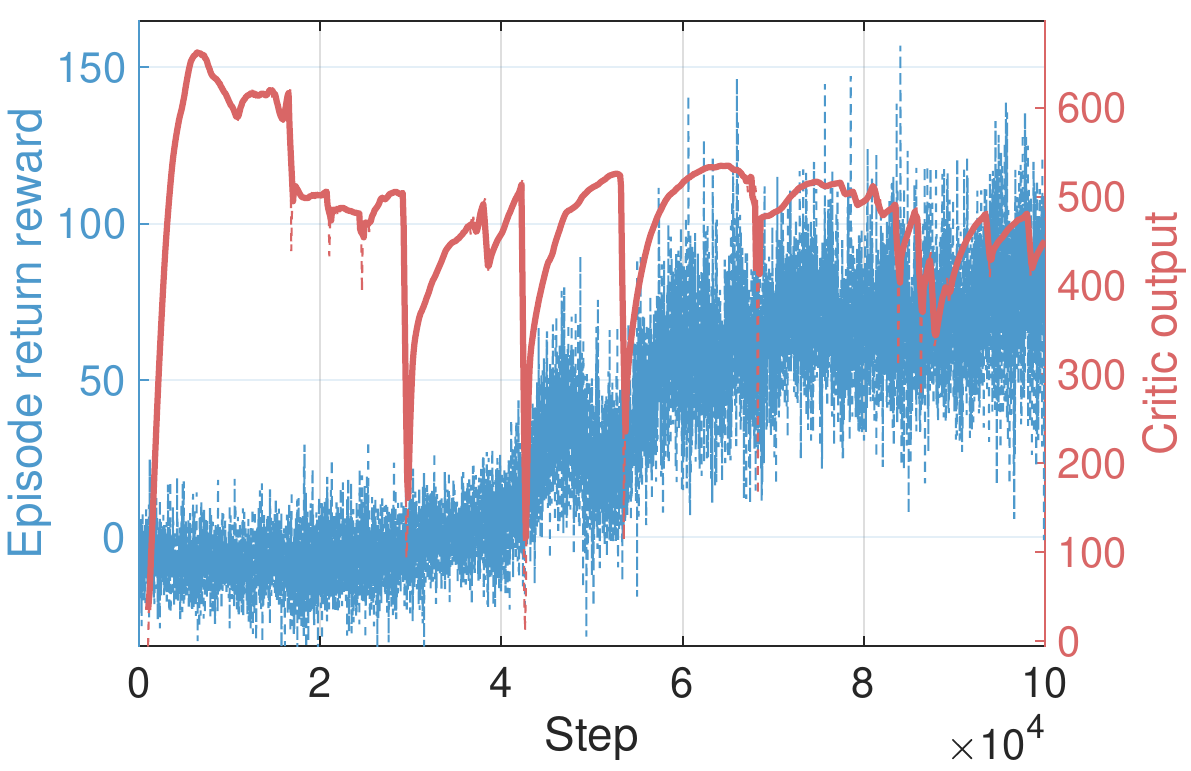}
}
\subfigure[HRea step number is 4, VRea step number is 3.]
{
\includegraphics[width=0.23\textwidth, height=0.18\textwidth]{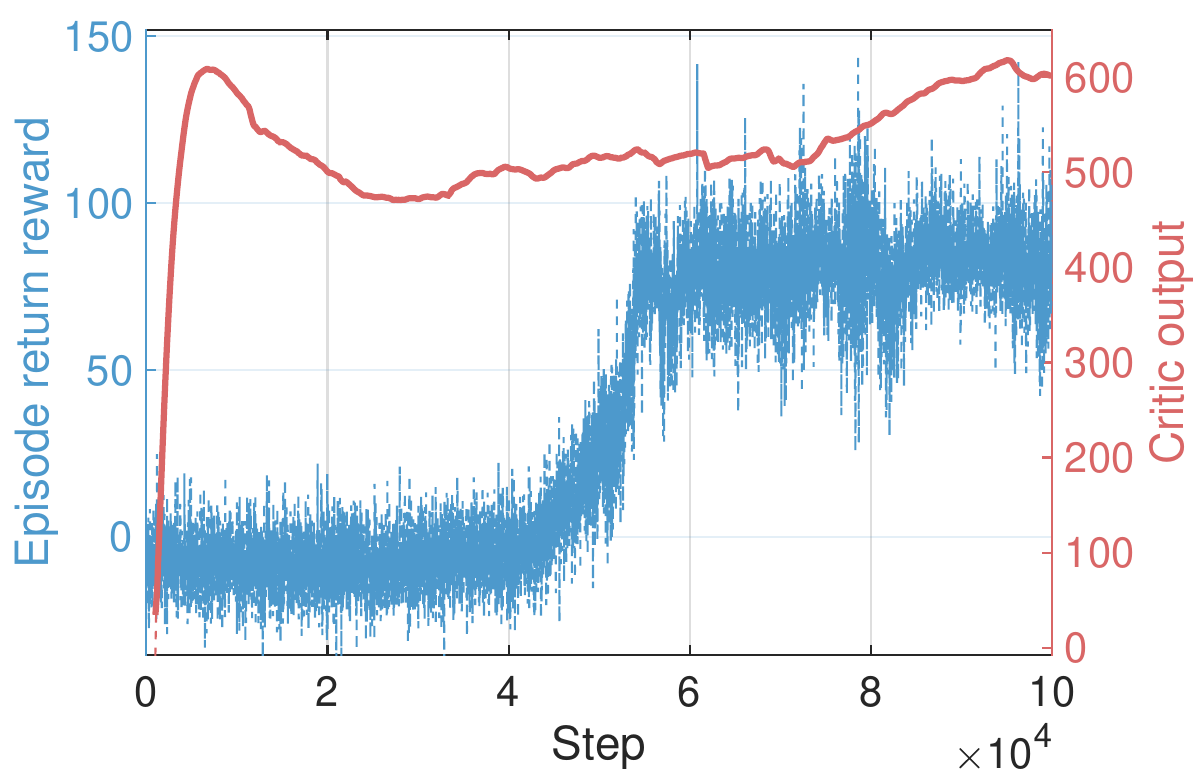}
}
\subfigure[HRea step number is 8, VRea step number is 1.]
{
\includegraphics[width=0.23\textwidth, height=0.18\textwidth]{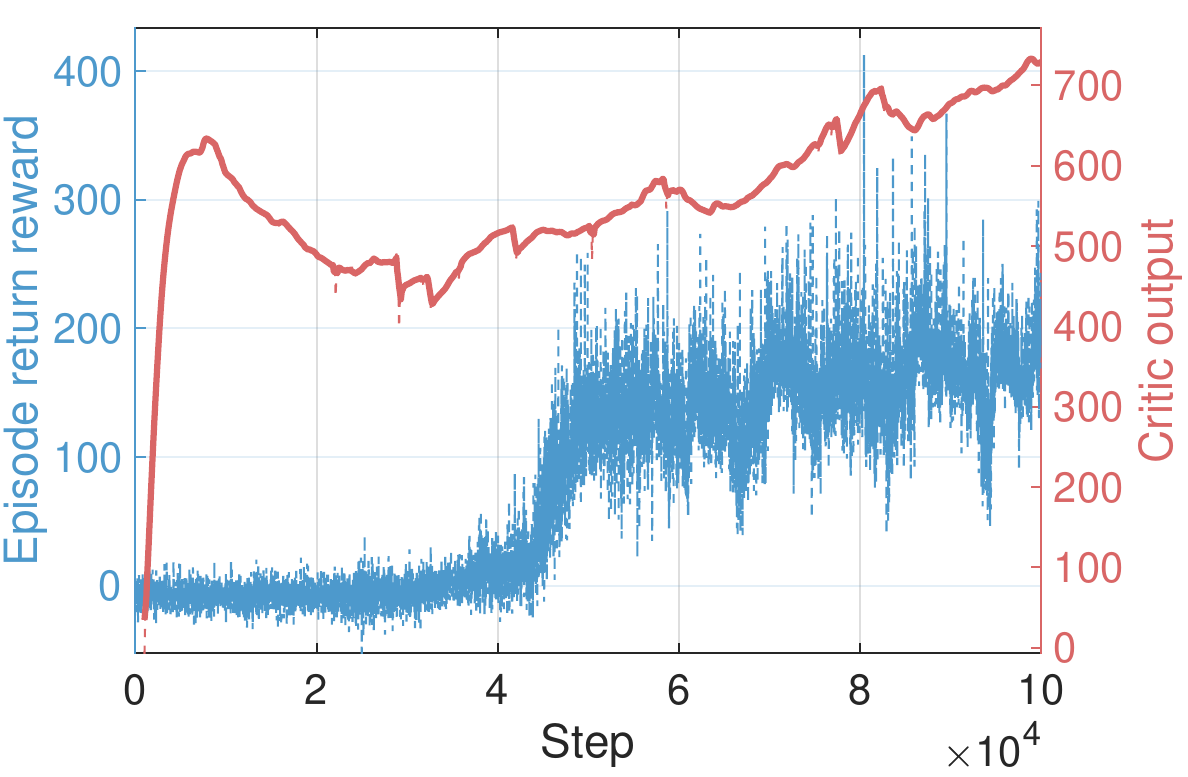}
}
\subfigure[HRea step number is 8, VRea step number is 3.]
{
\includegraphics[width=0.23\textwidth, height=0.18\textwidth]{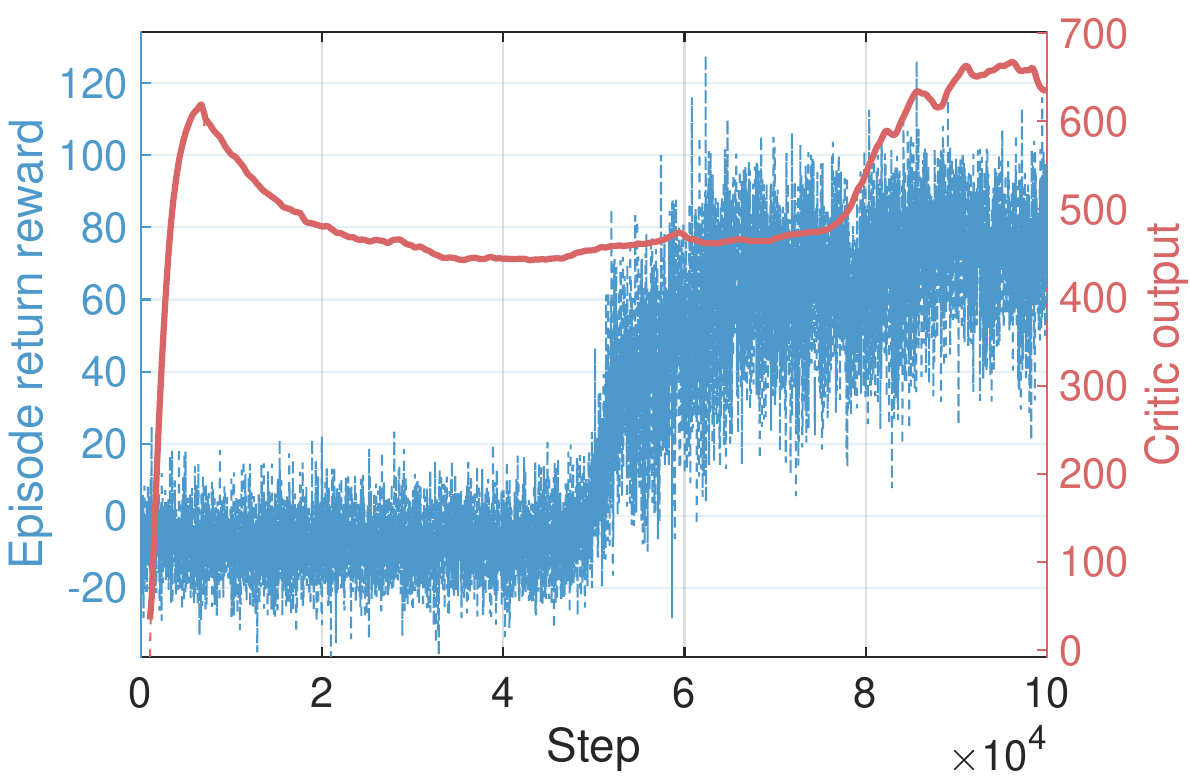}
}
\subfigure[HRea step number is 12, VRea step number is 3.]
{
\includegraphics[width=0.23\textwidth, height=0.18\textwidth]{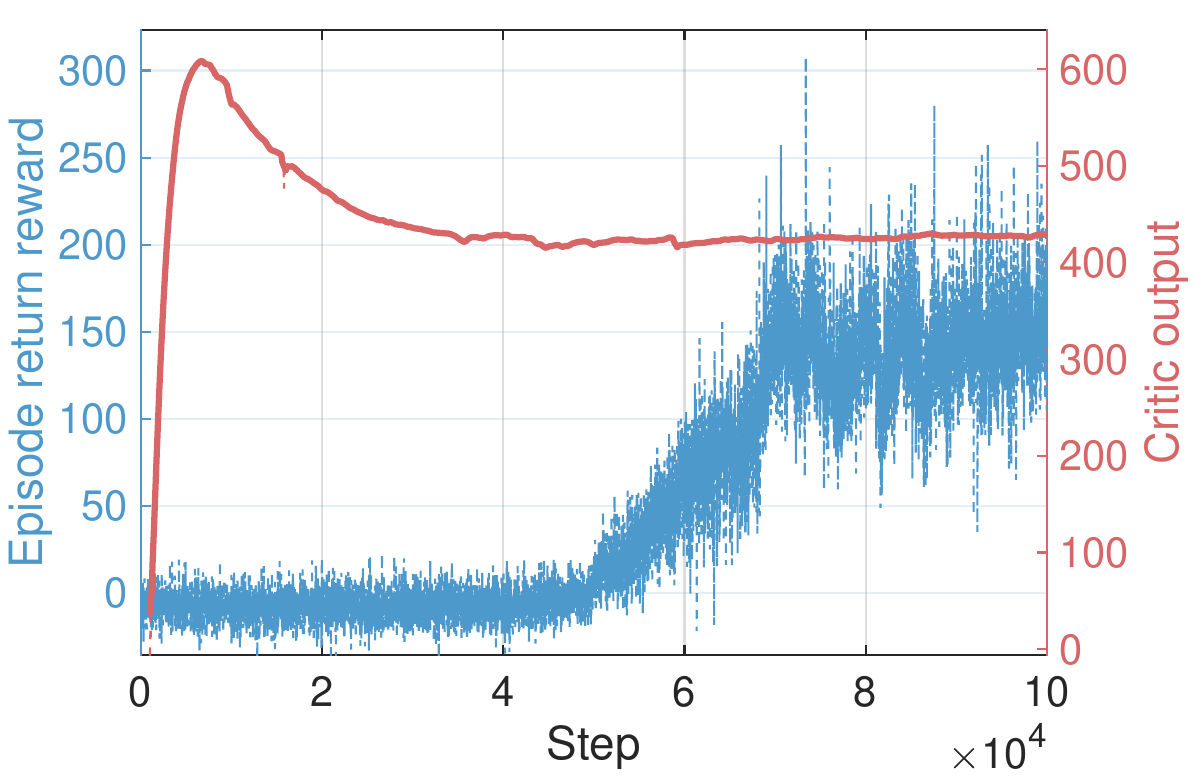}
}
\subfigure[HRea step number is 20, VRea step number is 1.]
{
\includegraphics[width=0.23\textwidth, height=0.18\textwidth]{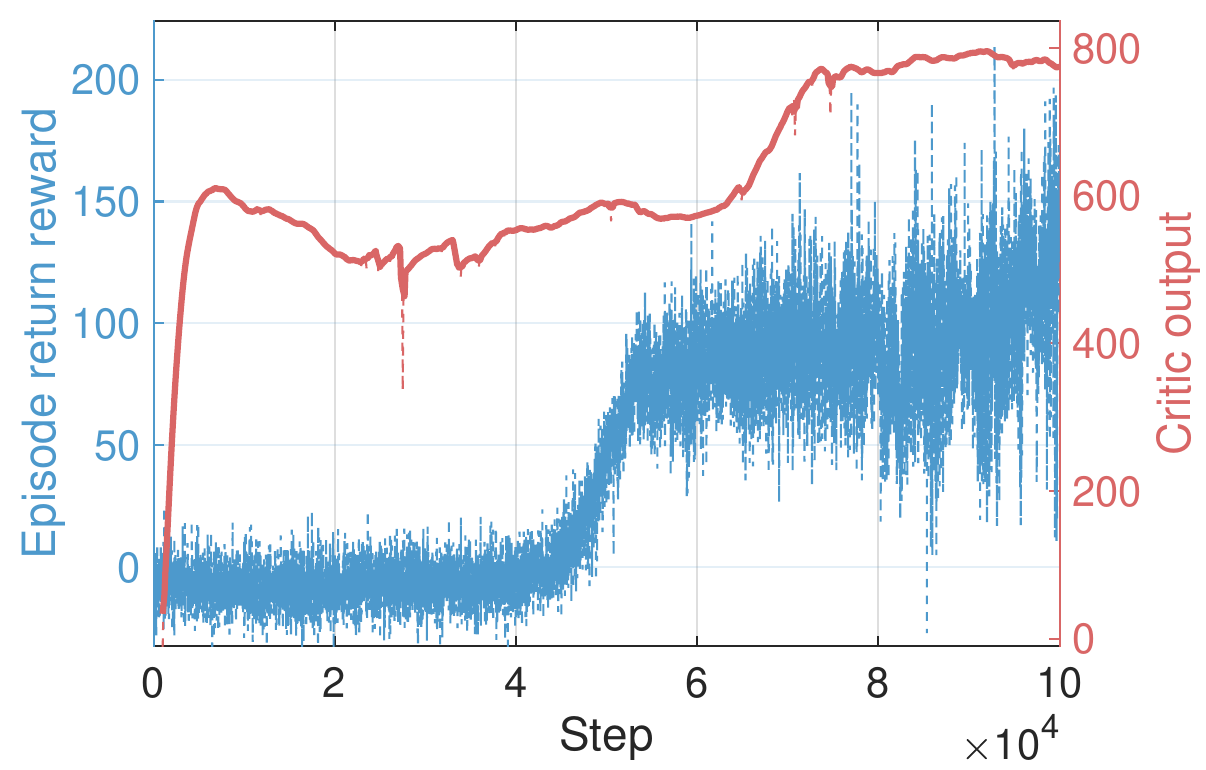}
}
\subfigure[SAC.]
{
\includegraphics[width=0.23\textwidth, height=0.18\textwidth]{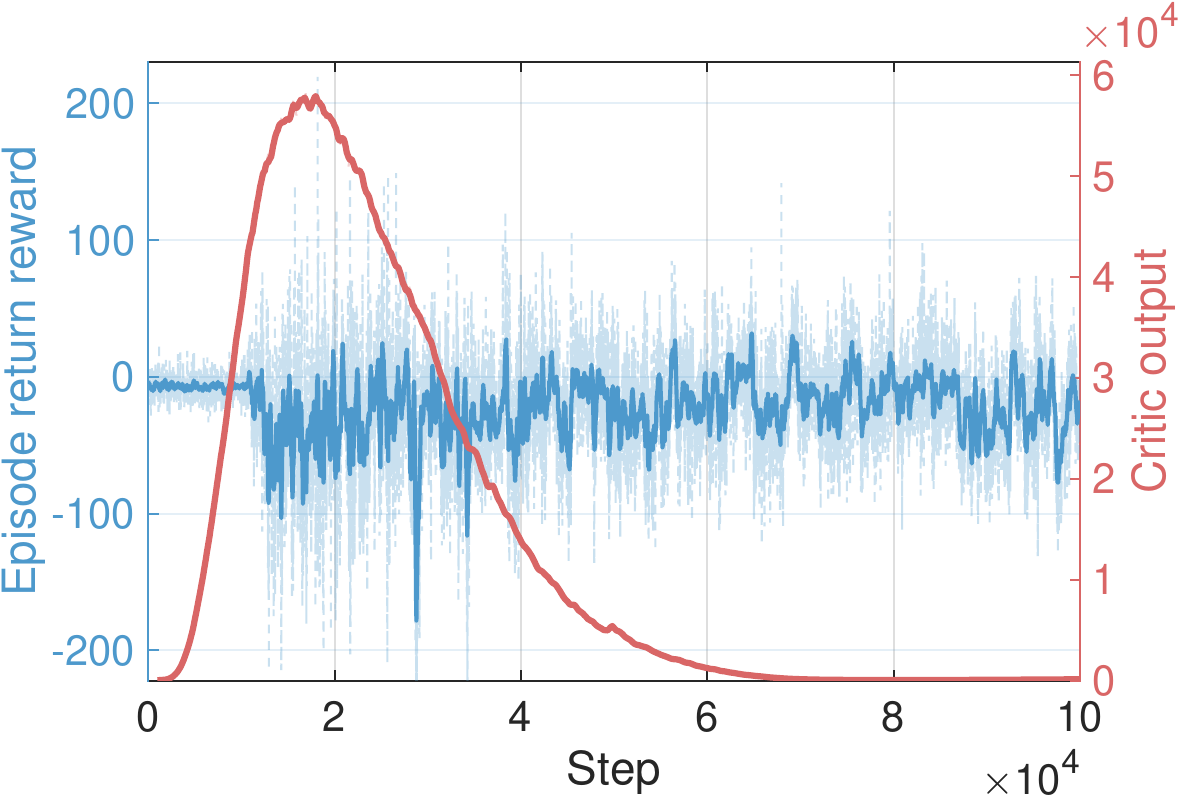}
}
\subfigure[Impact of HRea and VRea step settings on final reward in {\textit{ReaCritic}}-based SAC.]
{
\includegraphics[width=0.23\textwidth, height=0.18\textwidth]{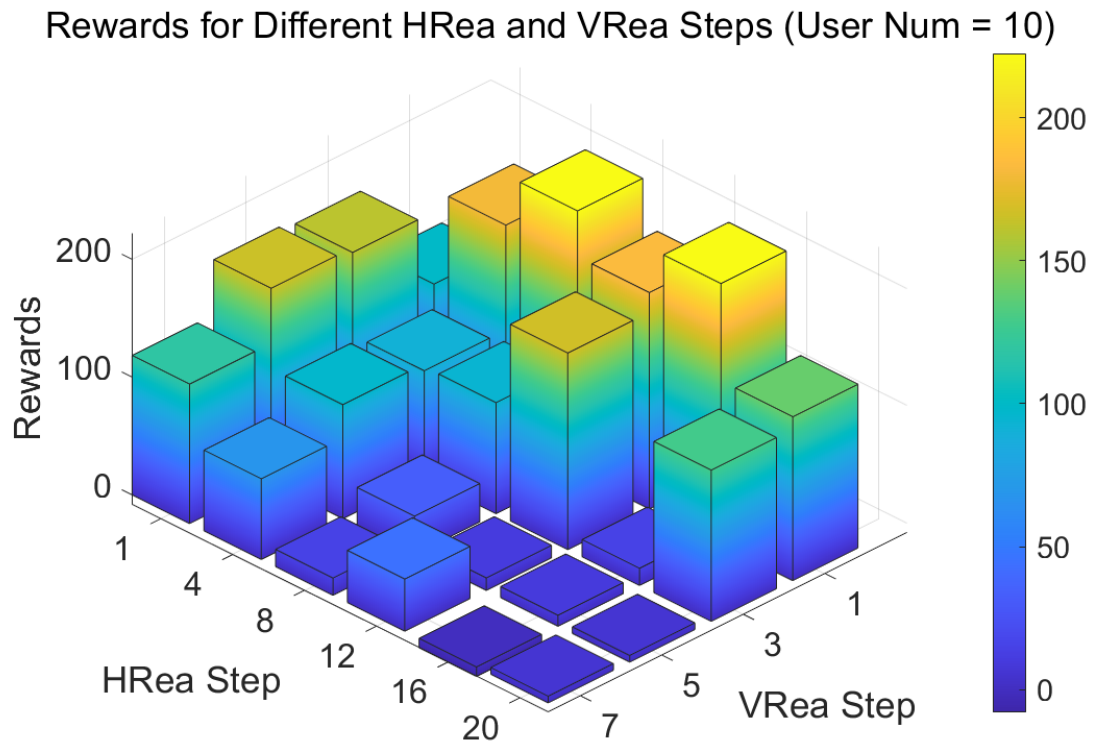}
}
\caption{Comparison of training performance with block number to be $4$, and  user number  to be 10 across {\textit{ReaCritic}}-based SAC method with varying numbers of HRea steps and VRea steps, and standard SAC. The final reward {\textit{ReaCritic}}-based SAC with different settings of HRea step and VRea step for 10 users is also included. } 
\label{fig:user10}
\end{figure*}

\begin{figure*}[t]
\centering
\subfigure[HRea step number is 1, VRea step number is 7.]
{
\includegraphics[width=0.23\textwidth, height=0.18\textwidth]{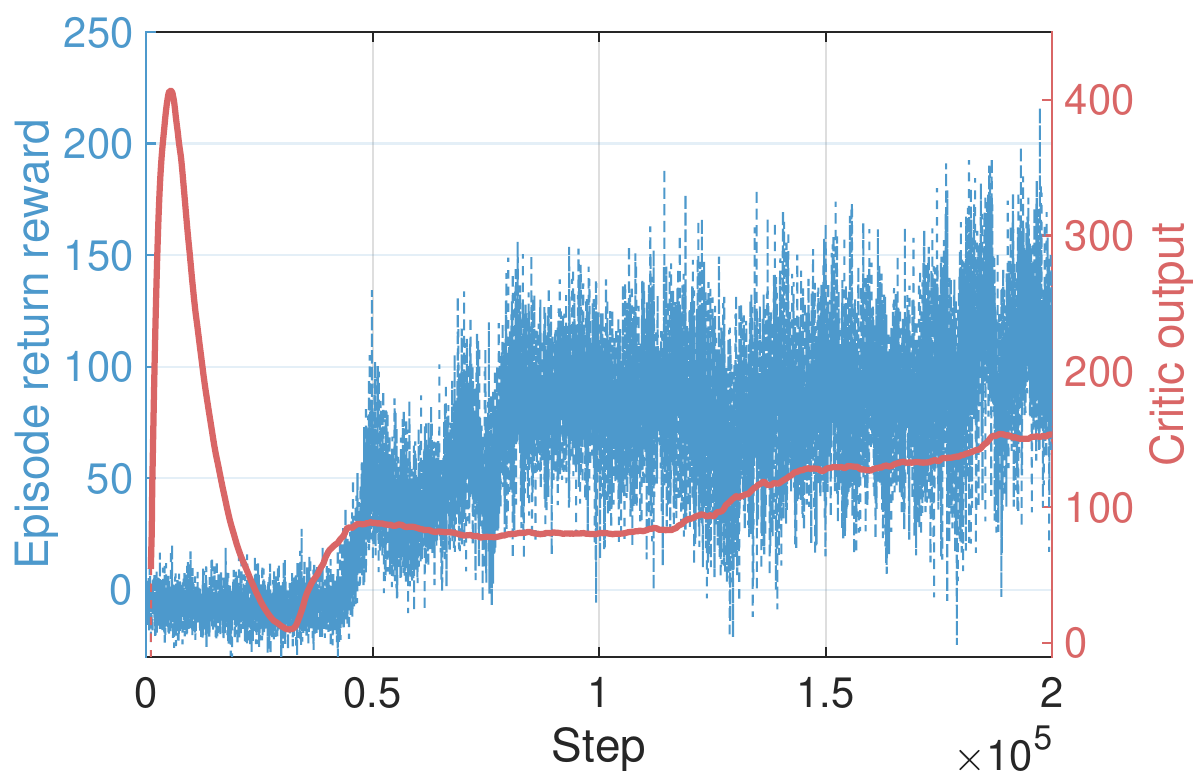}
}
\subfigure[HRea step number is 4, VRea step number is 5]
{
\includegraphics[width=0.23\textwidth, height=0.18\textwidth]{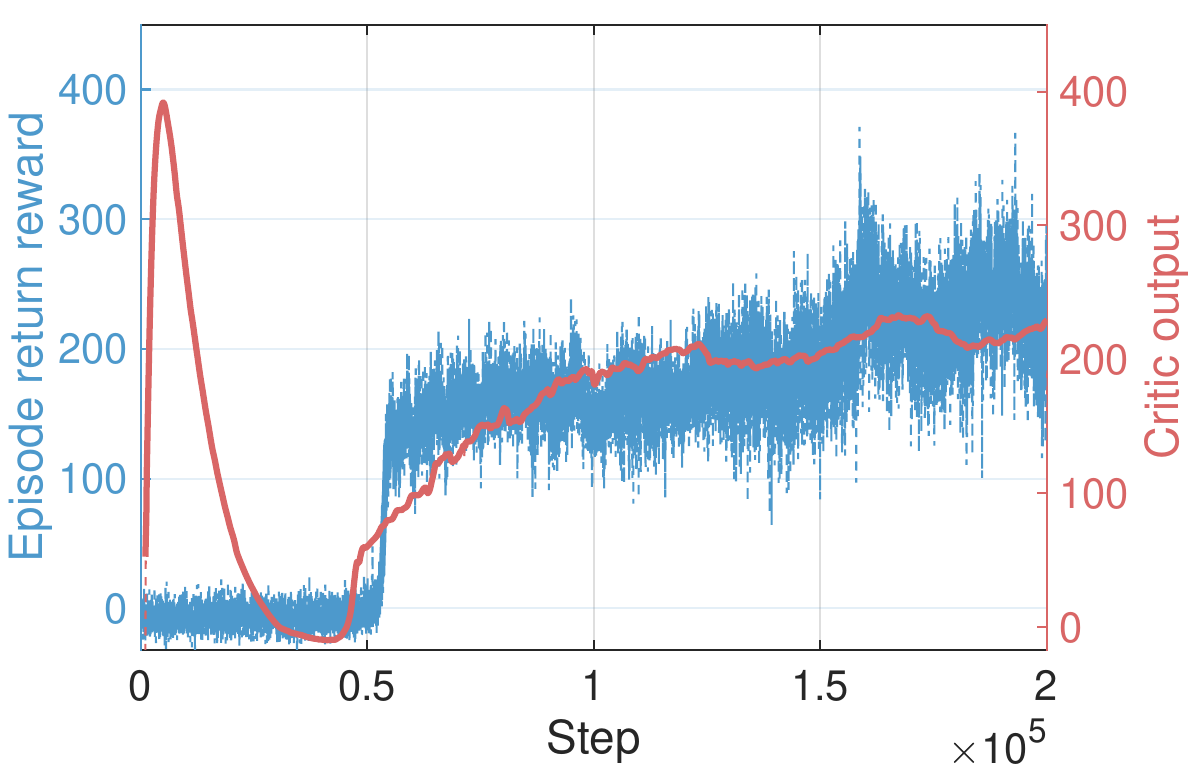}
}
\subfigure[HRea step number is 8, VRea step number is 1.]
{
\includegraphics[width=0.23\textwidth, height=0.18\textwidth]{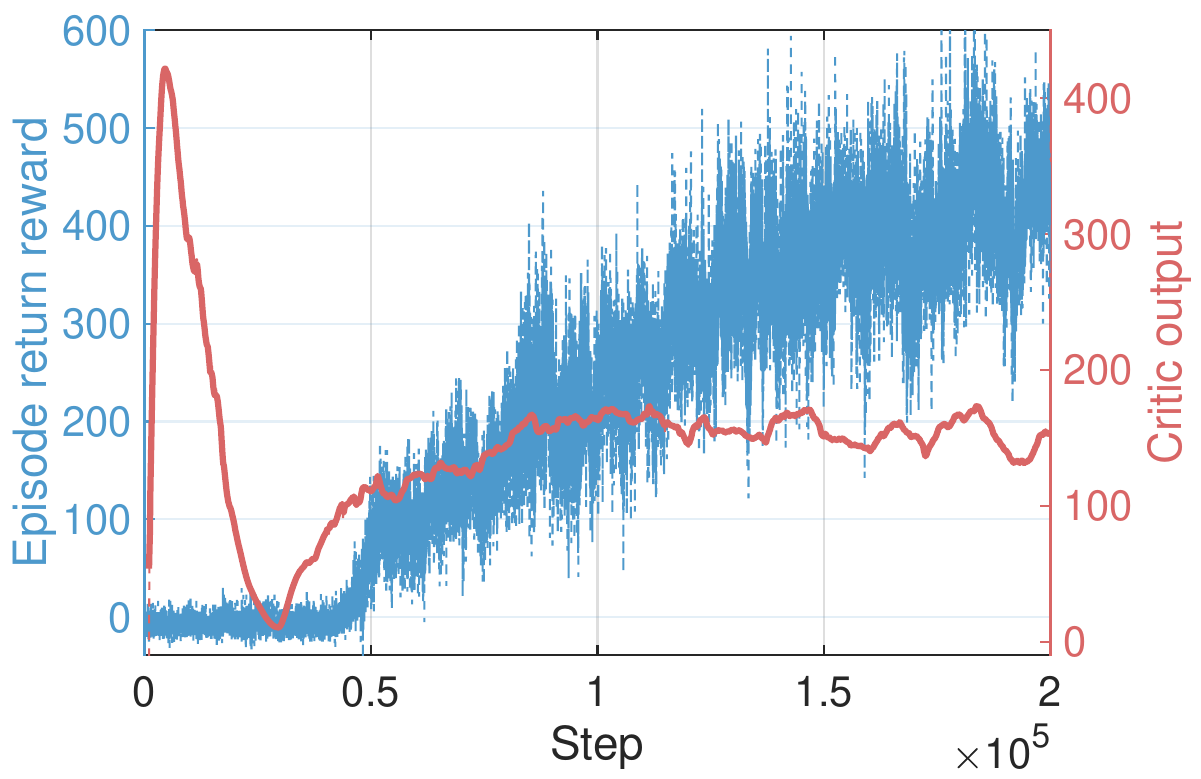}
}
\subfigure[HRea step number is 8, VRea step number is 3.]
{
\includegraphics[width=0.23\textwidth, height=0.18\textwidth]{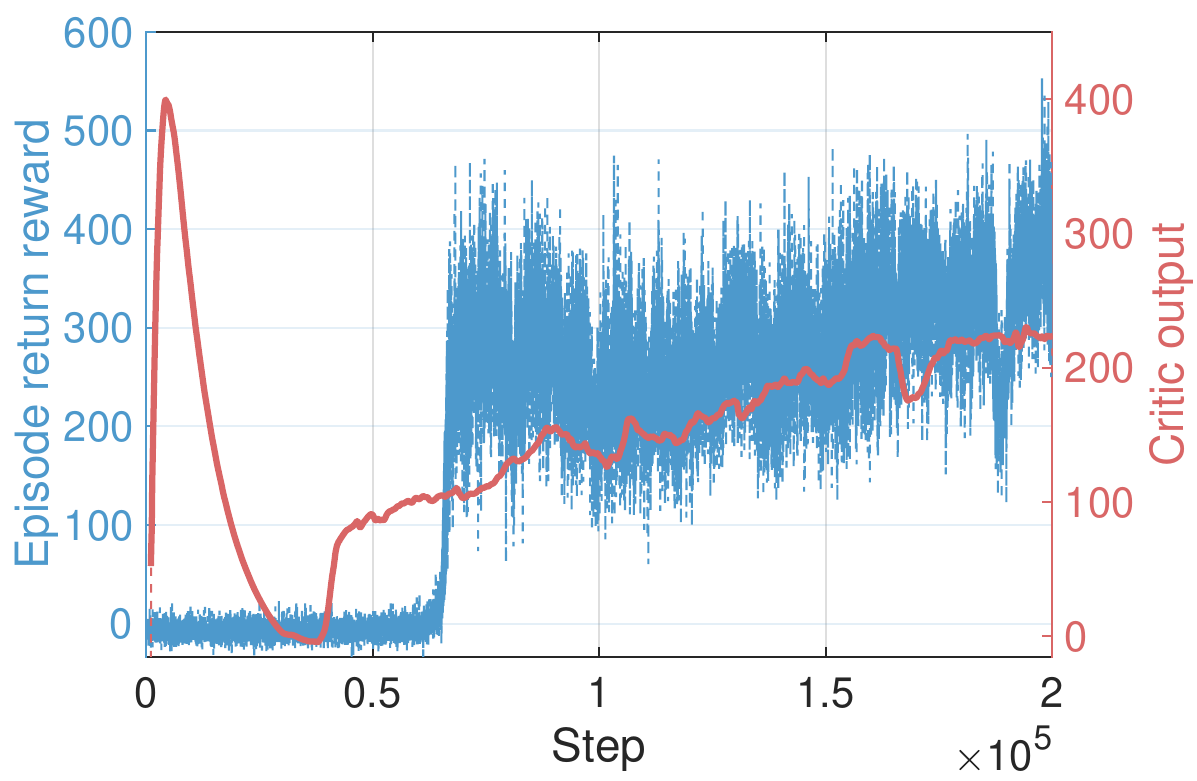}
}
\subfigure[HRea step number is 16, VRea step number is 3.]
{
\includegraphics[width=0.23\textwidth, height=0.18\textwidth]{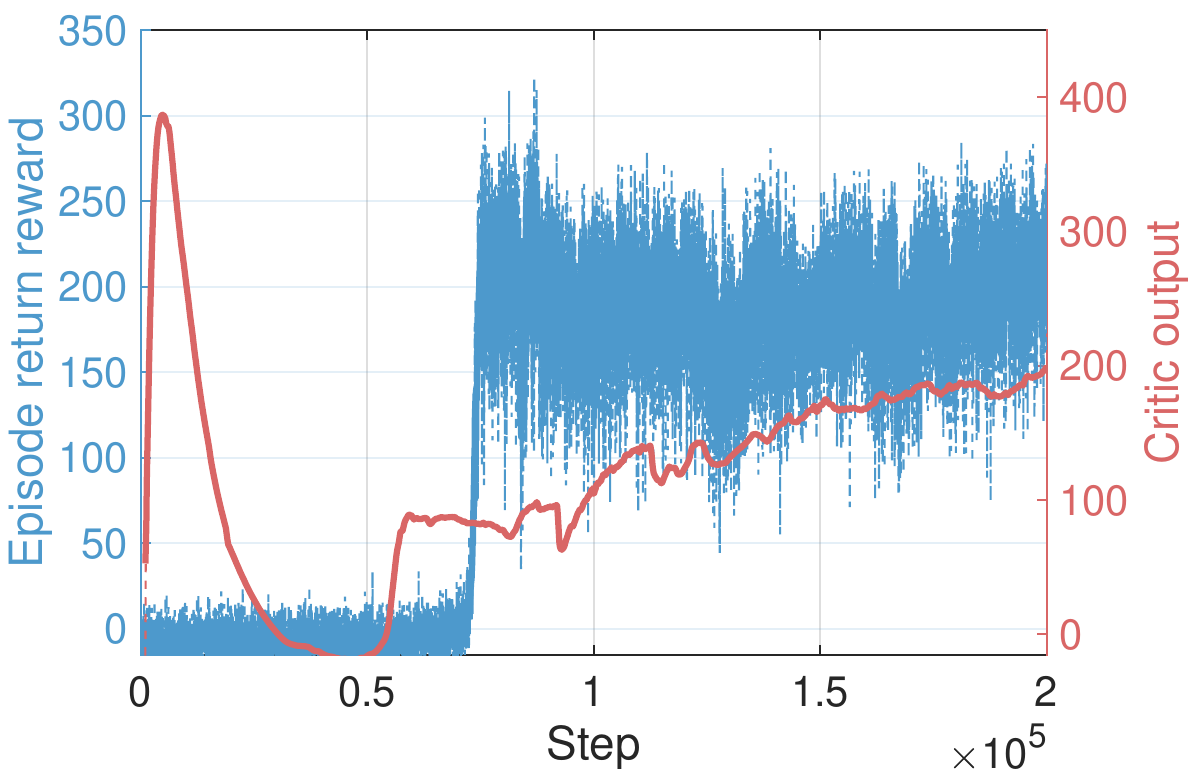}
}
\subfigure[HRea step number is 20, VRea step number is 1.]
{
\includegraphics[width=0.23\textwidth, height=0.18\textwidth]{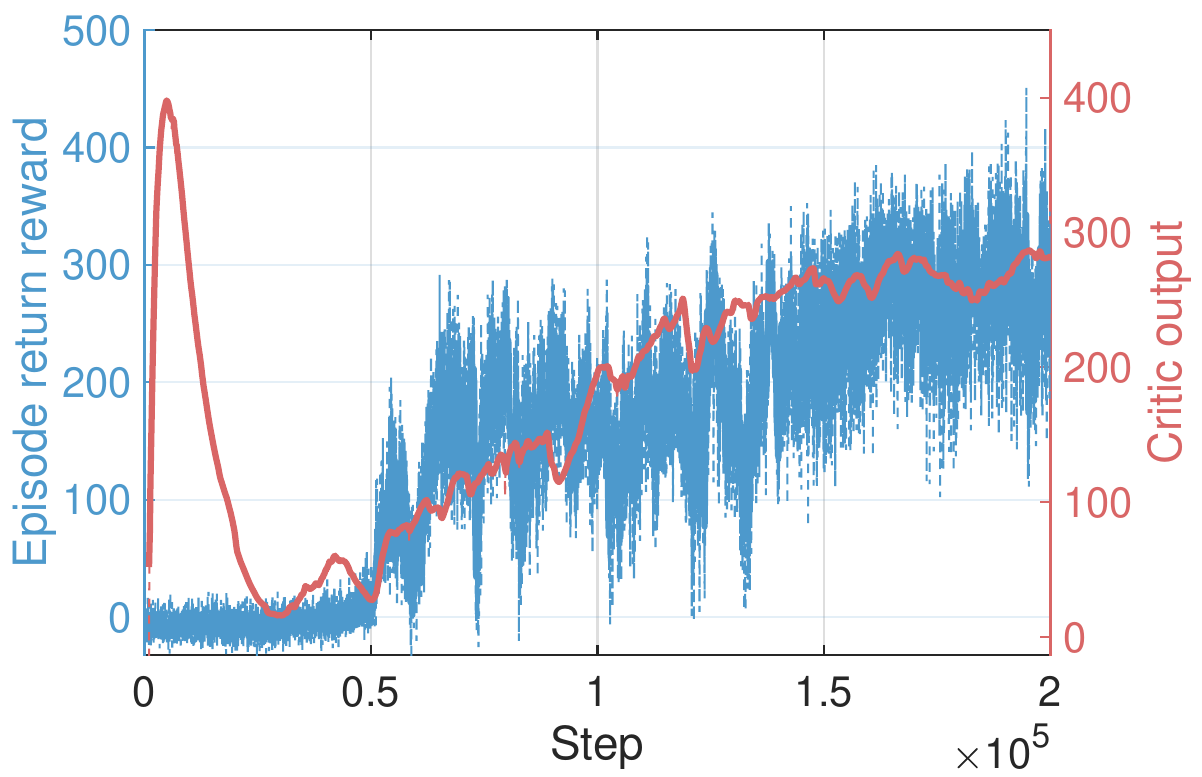}
}
\subfigure[SAC.]
{
\includegraphics[width=0.23\textwidth, height=0.18\textwidth]{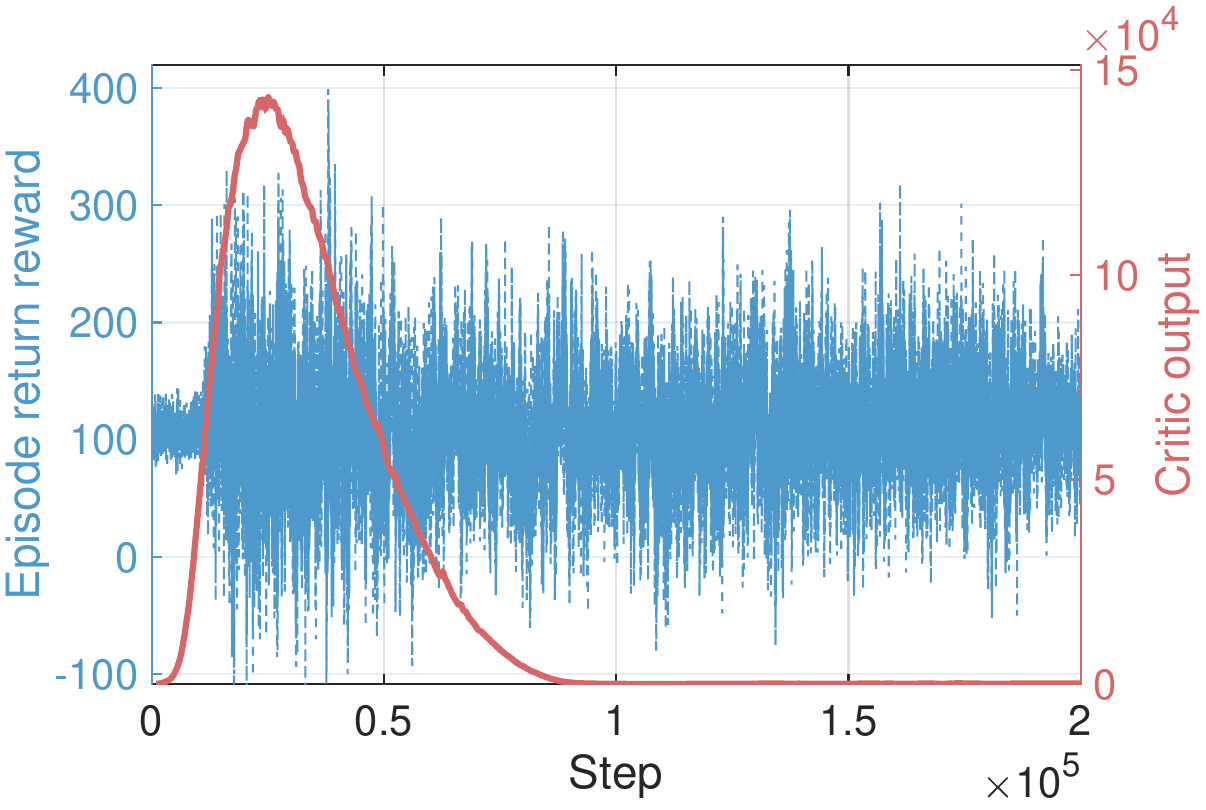}
}
\subfigure[Impact of HRea and VRea step settings on final reward in {\textit{ReaCritic}}-based SAC with 20 users.]
{
\includegraphics[width=0.23\textwidth, height=0.18\textwidth]{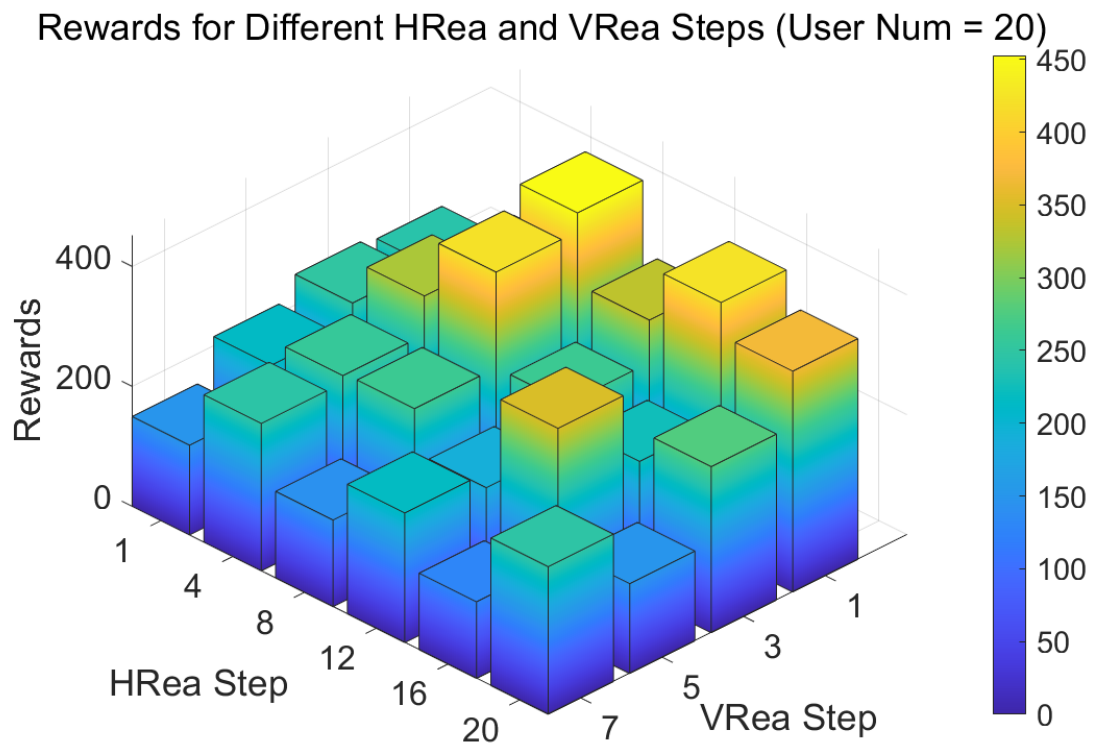}
}
\caption{Comparison of training performance with block number to be $4$, and  user number  to be 20 across {\textit{ReaCritic}}-based SAC method, and standard SAC. The final reward {\textit{ReaCritic}}-based SAC with different settings of HRea step and VRea step for 20 users is also included.}
\label{fig:user20}
\end{figure*}

\begin{figure*}[t]
\centering
\subfigure[HRea step number is 4, VRea step number is 7.]
{
\includegraphics[width=0.23\textwidth, height=0.18\textwidth]{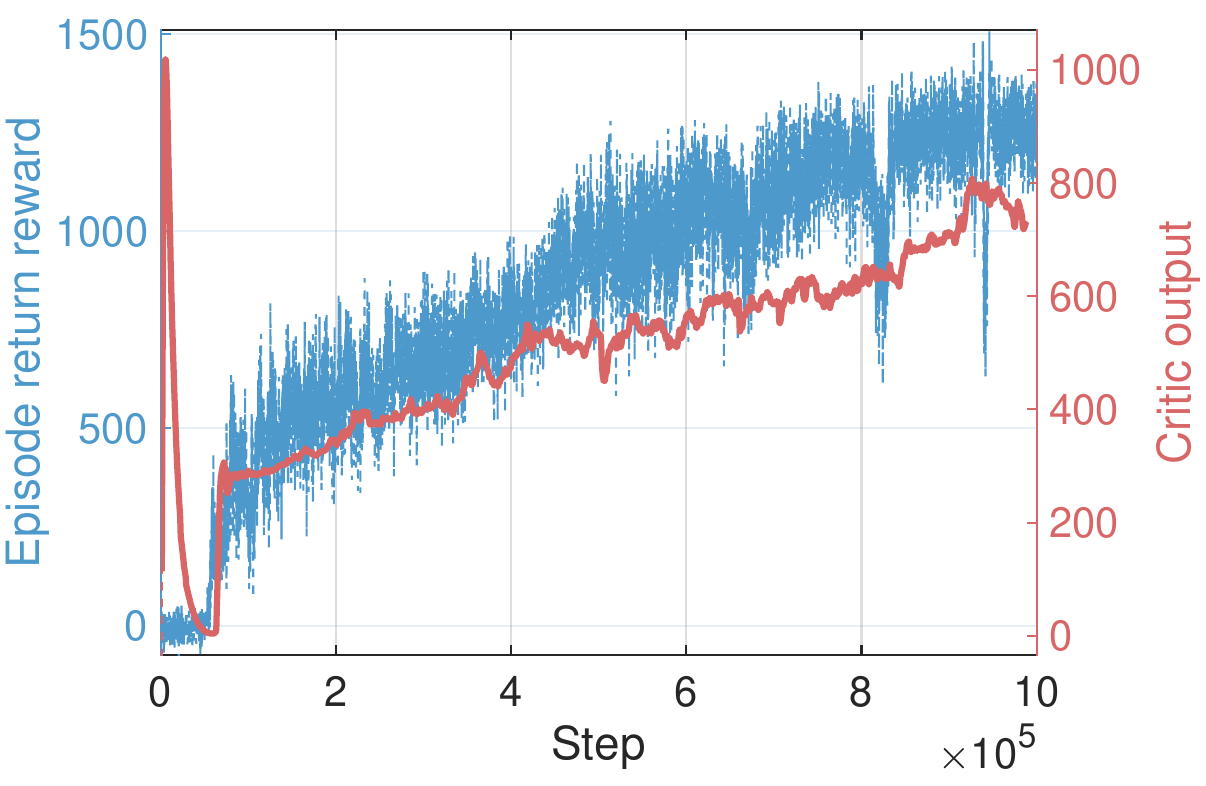}
}
\subfigure[HRea step number is 8, VRea step number is 1.]
{
\includegraphics[width=0.23\textwidth, height=0.18\textwidth]{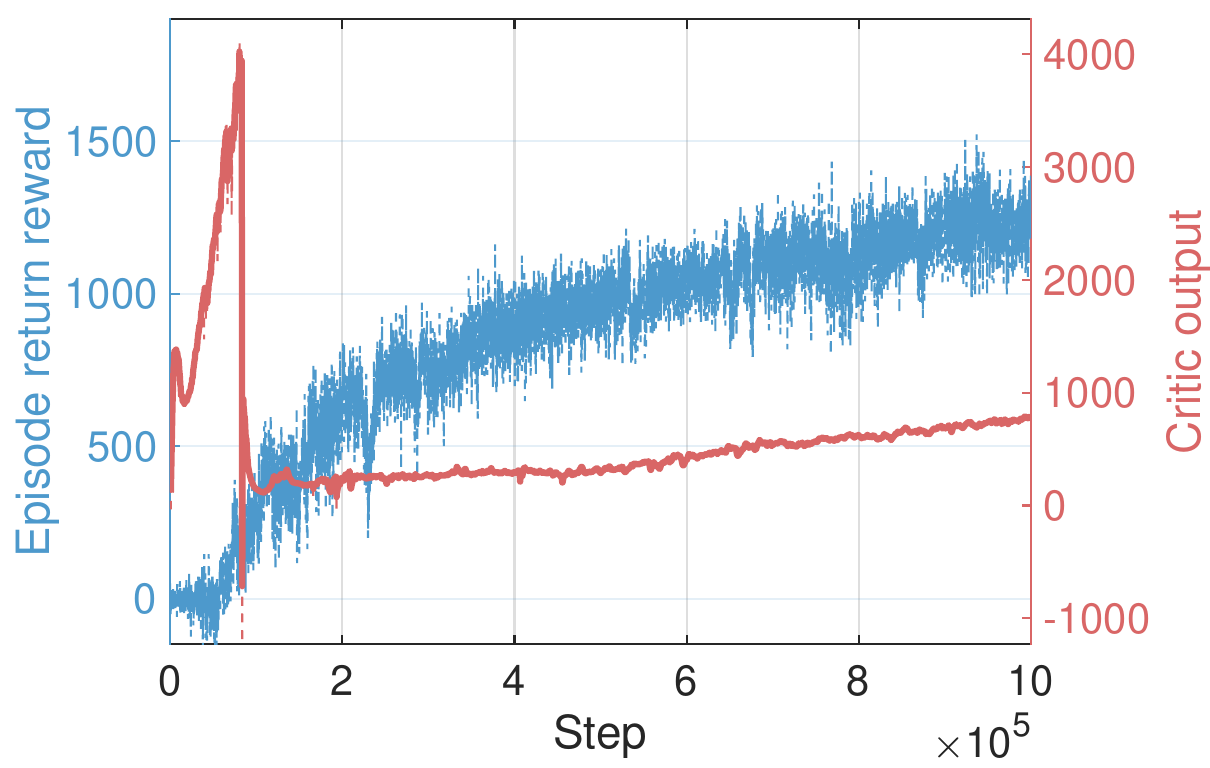}
}
\subfigure[HRea step number is 8, VRea step number is 3.]
{
\includegraphics[width=0.23\textwidth, height=0.18\textwidth]{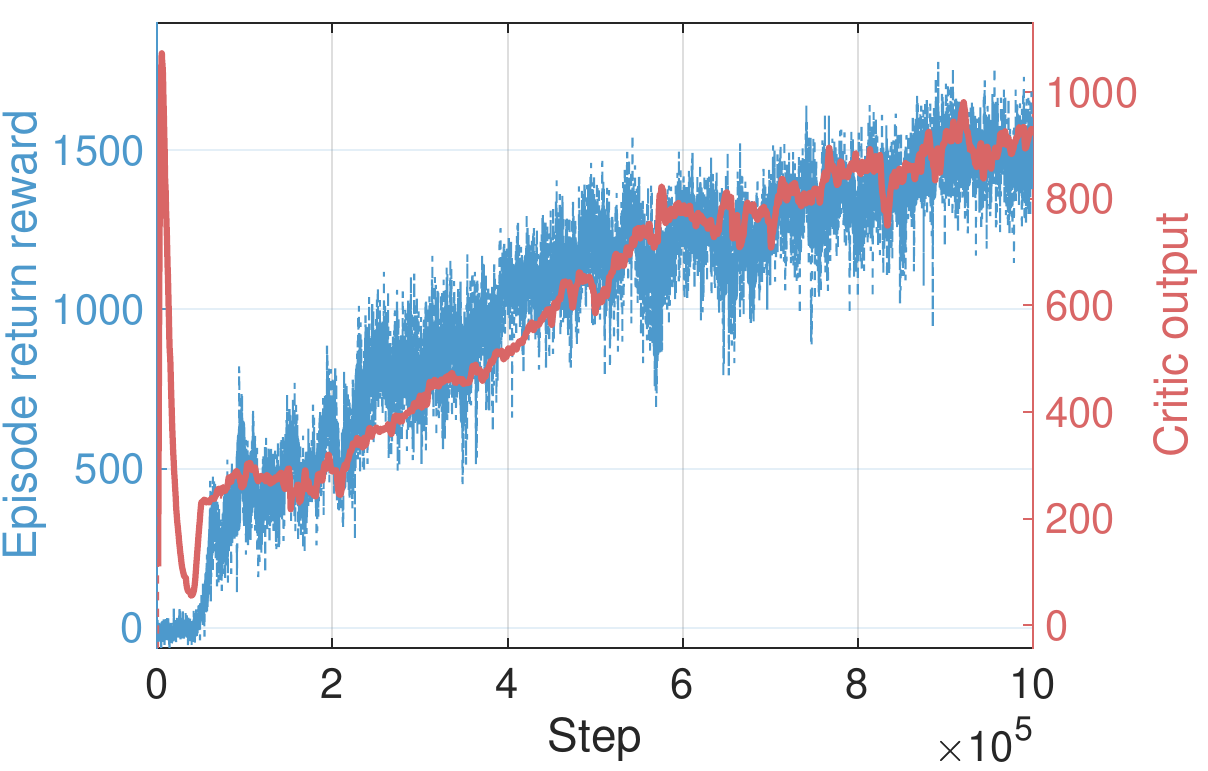}
}
\subfigure[HRea step number is 12, VRea step number is 3.]
{
\includegraphics[width=0.23\textwidth, height=0.18\textwidth]{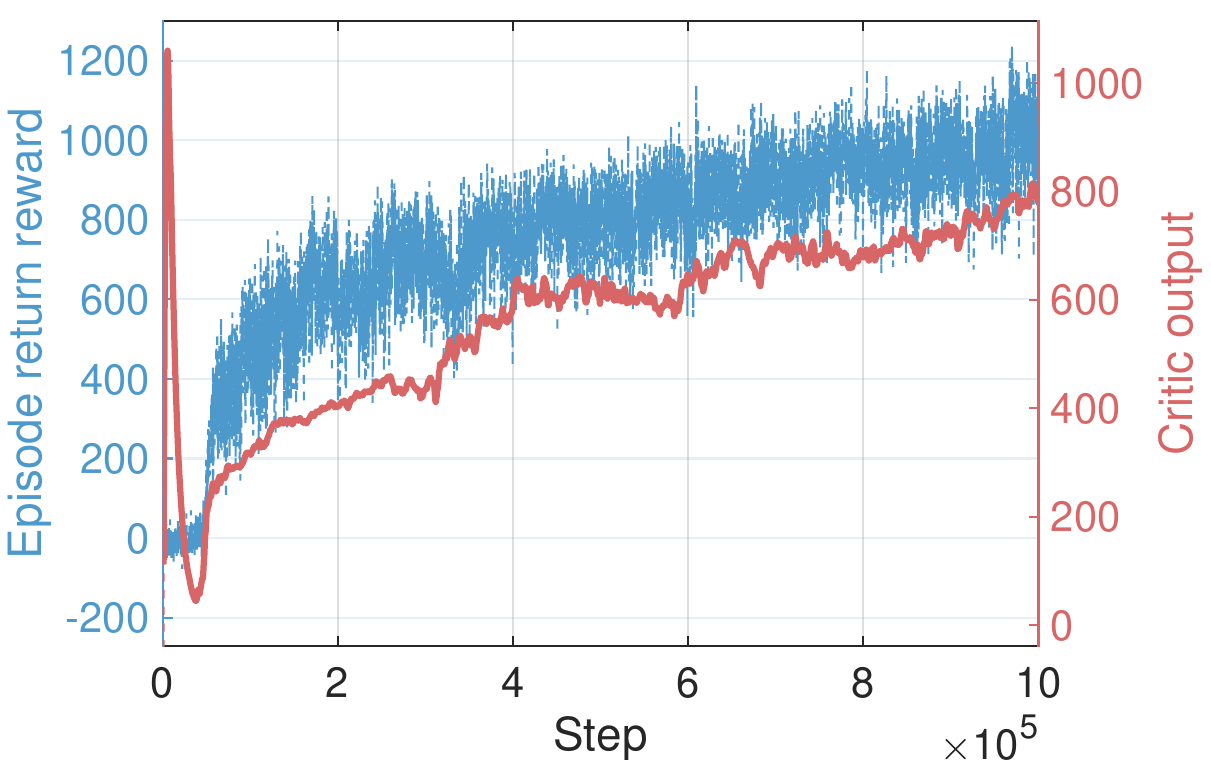}
}
\subfigure[HRea step number is 16, VRea step number is 1.]
{
\includegraphics[width=0.23\textwidth, height=0.18\textwidth]{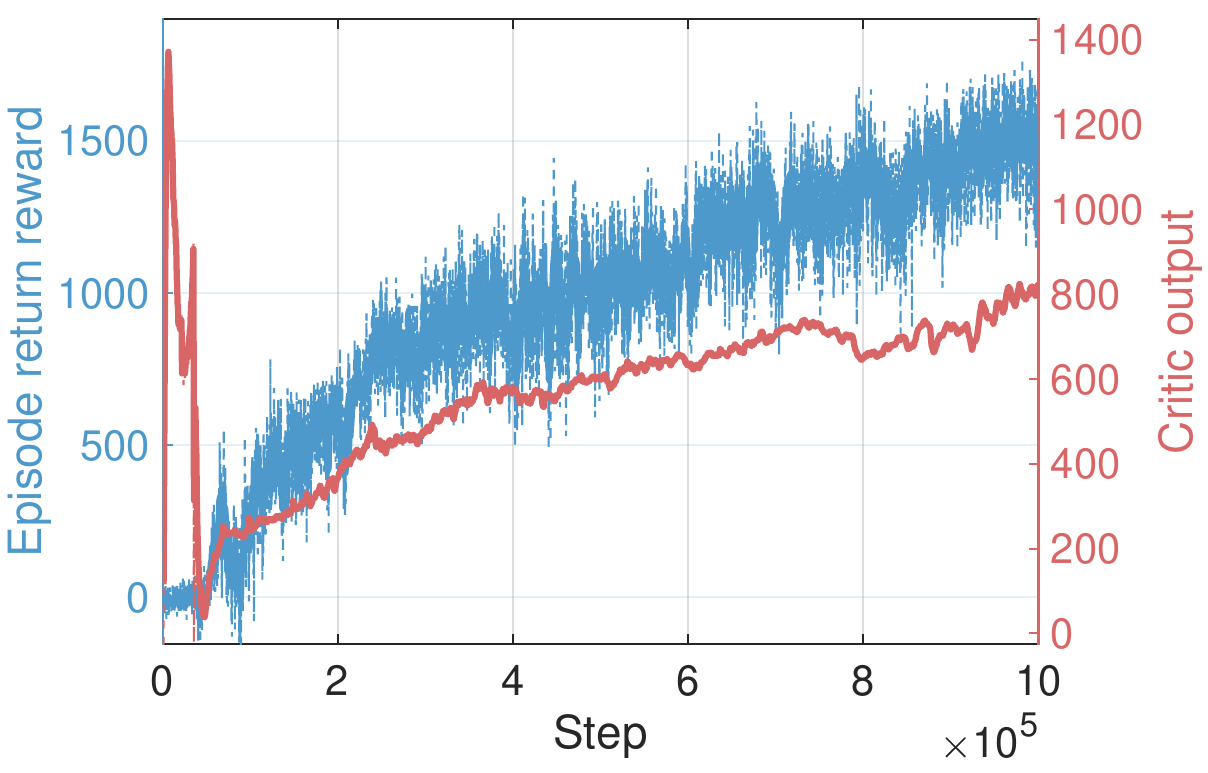}
}
\subfigure[HRea step number is 20, VRea step number is 1.]
{
\includegraphics[width=0.23\textwidth, height=0.18\textwidth]{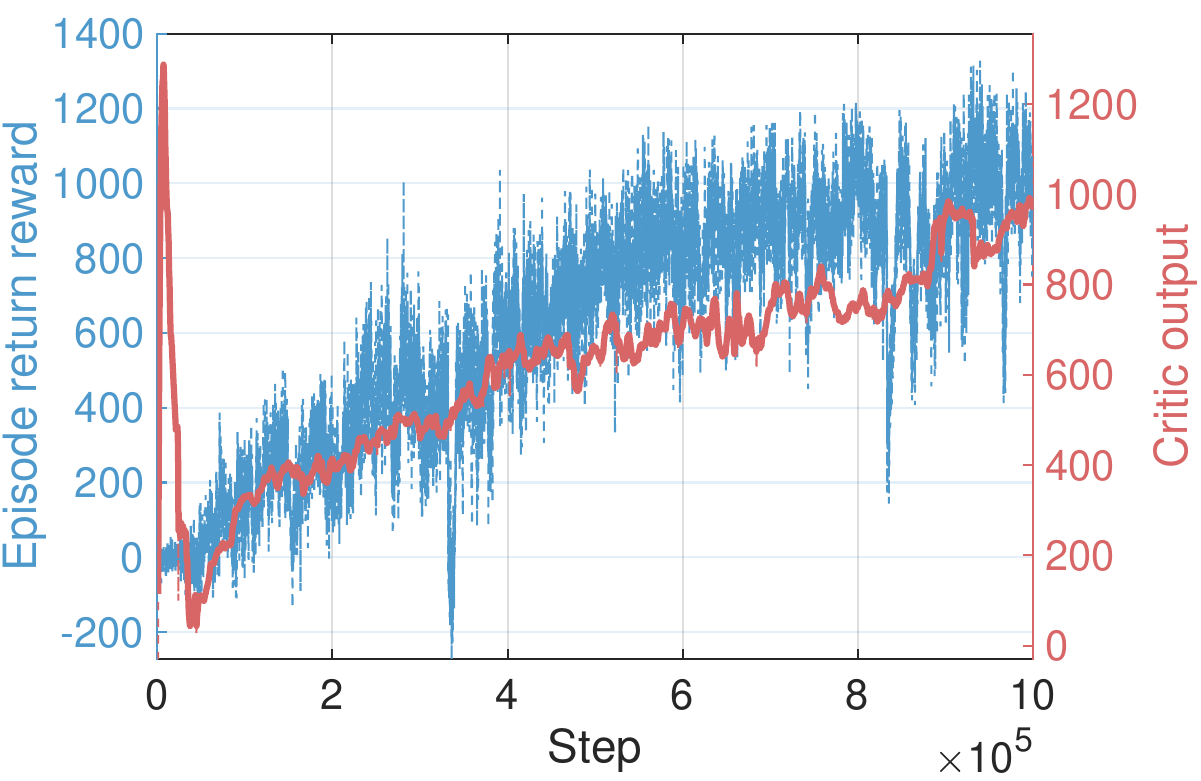}
}
\subfigure[SAC.]
{
\includegraphics[width=0.23\textwidth, height=0.18\textwidth]{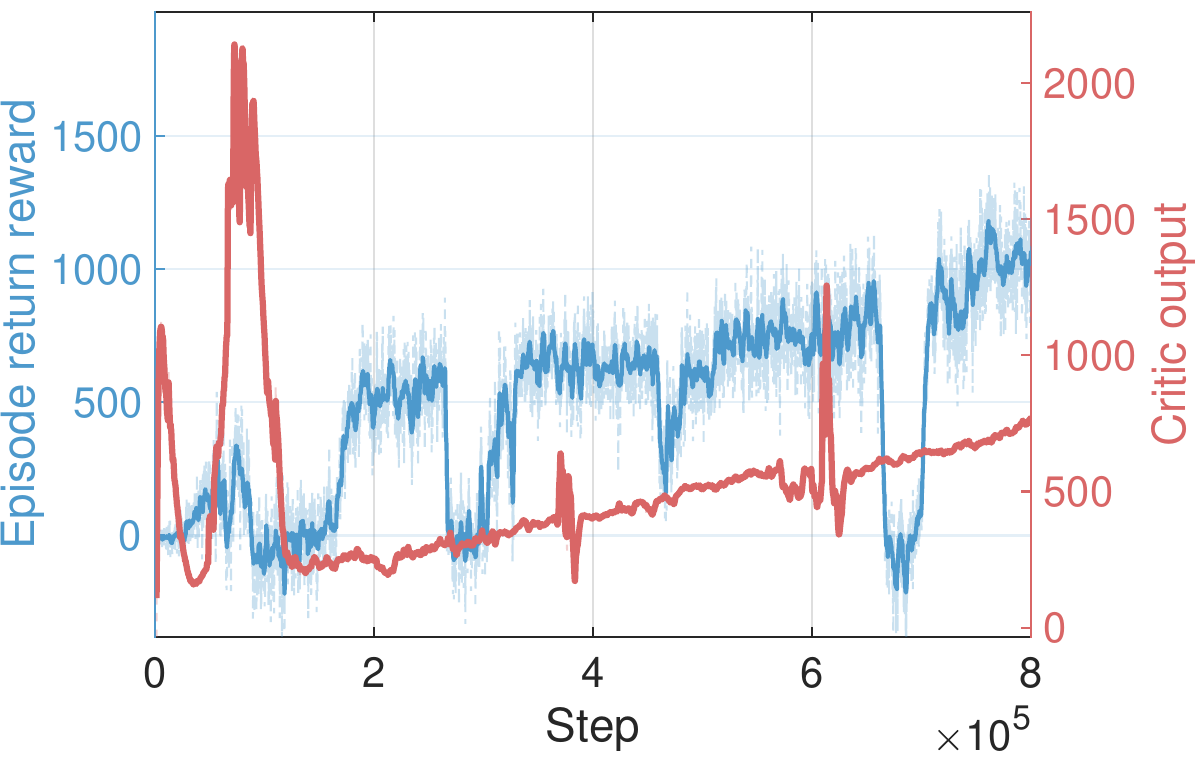}
}
\subfigure[Impact of HRea and VRea step settings on final reward in {\textit{ReaCritic}}-based SAC with 50 users.]
{
\includegraphics[width=0.23\textwidth, height=0.18\textwidth]{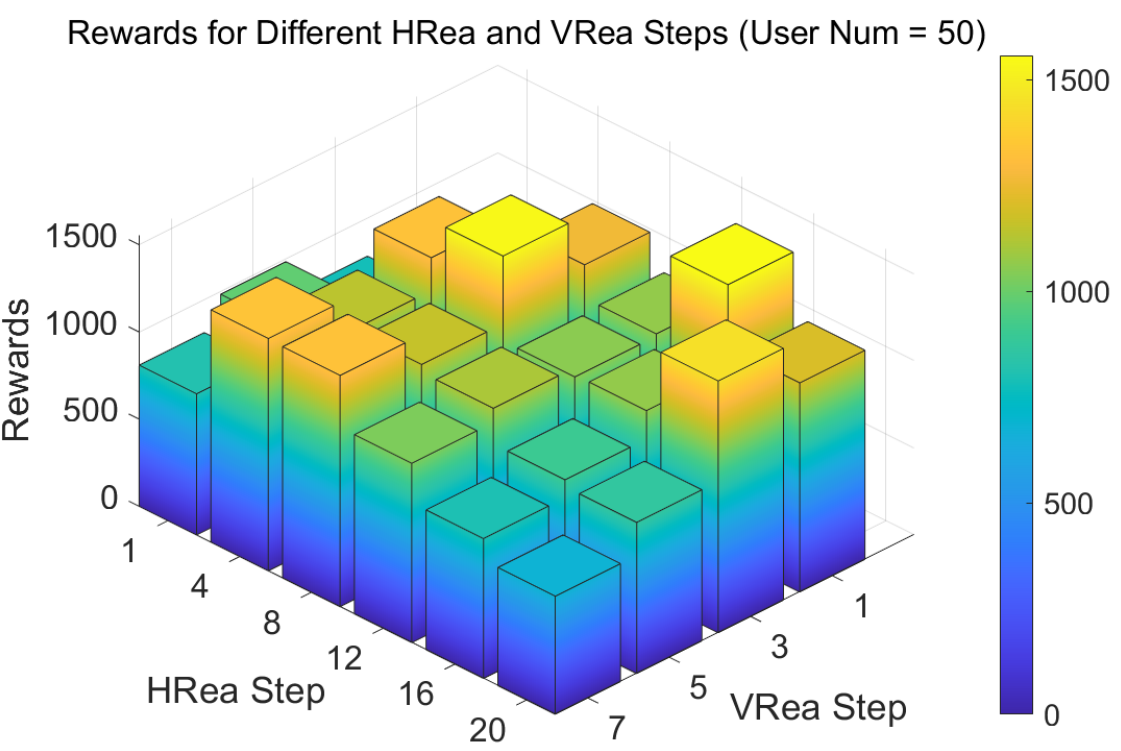}
}
\caption{Comparison of training performance with block number to be $4$, and  user number  to be 50 across {\textit{ReaCritic}}-based SAC method, and standard SAC.  The final reward {\textit{ReaCritic}}-based SAC with different settings of HRea step and VRea step for 50 users is also included.}
\label{fig:user50}
\end{figure*}
Fig.~\ref{fig:user10} compares the training performance of ReaCritic-based SAC under different HRea and VRea configurations with a small HetNet of $10$ users. \textcolor{black}{Notably, the configuration $(H=1, V=1)$ illustrated in Fig.~\ref{fig:user10}(a) serves as the foundational Transformer-based baseline. This setting represents a vanilla Transformer-based critic that utilizes self-attention to capture user dependencies but lacks the iterative reasoning-time scaling proposed in this work.} Compared to the standard SAC baseline, ReaCritic consistently achieves higher and more stable cumulative rewards. \textcolor{black}{The relationship between HRea and VRea is complementary: while HRea steps expand the reasoning breadth by generating multiple parallel tokens to explore state-action correlations, VRea steps enhance the reasoning depth through hierarchical abstraction. The results show that although the $(1,1)$ baseline outperforms standard MLP-based SAC, scaling both dimensions to configurations such as $(4,3)$ or $(8,1)$ remains necessary to fully capture the intricate coupling in wireless networks.} When both HRea and VRea steps are small, e.g., $(4,3)$ or $(8,1)$, the model already outperforms standard SAC. Moreover, the critic output remains stable and well-aligned with the reward trajectory, indicating a coherent value function approximation. In contrast, standard SAC exhibits unstable critic behavior, with initial improvement followed by rapid collapse. This weakens the learning signal and degrades policy performance.
To analyze scalability, we evaluate medium-sized ($M=20$) and large-scale ($M=50$) HetNets in Fig.~\ref{fig:user20} and Fig.~\ref{fig:user50}, respectively. As the number of users increases, the HetNet becomes more heterogeneous and the state-action space grows substantially, leading to a more complex learning environment. The results show that higher HRea and VRea settings, e.g., $(12,3)$, $(16,5)$, yield better final rewards in such settings. \textcolor{black}{The results confirm that while the basic $(1,1)$ Transformer baseline provides a stable starting point, higher configurations like $(12,3)$ or $(16,5)$ are vital for handling the increased interaction complexity in large-scale scenarios.} This validates our hypothesis that larger reasoning capacity, enabled by deeper horizontal and vertical reasoning, is needed to model the increased interaction complexity.

Across all user scales, ReaCritic-based SAC consistently outperforms standard SAC in both convergence speed and final performance. Furthermore, the Q-value curves in ReaCritic cases maintain stability throughout training, confirming that the structured inference paths constructed by the HRea and VRea mechanisms facilitate more reliable and expressive value estimation. These findings highlight ReaCritic’s scalability and robustness in DRL training for dynamic, multi-user HetNets.

\subsubsection{Generalization across DRL Algorithms and Benchmark Environments}
To verify the versatility of \textit{ReaCritic}, we integrate it with multiple DRL backbones (SAC, TD3, DDPG, PPO, A3C), and evaluate performance on three OpenAI Gym tasks with varying complexity: \textit{MountainCarContinuous-v0}, \textit{HumanoidStandup-v4}, and \textit{Ant-v4}.

\begin{table}[t]
\centering
\caption{Final Episodic Return on \textit{HumanoidStandup-v4} With and Without \textit{ReaCritic}}
\label{tab:rea_vs_baseline}
{\small \begin{tabular}{l|c|c|c}
\toprule
\textbf{Algorithm} & \textbf{With} & \textbf{Without} & \textbf{Gain} \\
\midrule
SAC  & 158.7k & 152.9k & +3.8\% \\
DDPG & 85.2k  & 31.5k  & +170.8\% \\
A3C  & 57.8k  & 53.9k  & +7.3\% \\
TD3  & 53.2k  & 42.1k  & +26.5\% \\
PPO  & 52.9k  & 51.0k  & +3.7\% \\
\bottomrule
\end{tabular}}
\end{table}

Table~\ref{tab:rea_vs_baseline} presents the final episodic return of five representative DRL algorithms on the \textit{HumanoidStandup-v4} task, comparing variants with and without \textit{ReaCritic}. Across all methods, integrating \textit{ReaCritic} leads to consistent performance improvements, with gains ranging from modest (e.g., SAC and PPO: +3.8\% and +3.7\%) to substantial (e.g., DDPG: +170.8\%). These results demonstrate \textit{ReaCritic}’s strong generalizability across different DRL backbones and task complexities. \textcolor{black}{These results demonstrate the robust generalizability of \textit{ReaCritic} across different DRL backbones, validating its efficacy as a versatile enhancement framework for complex, high-dimensional tasks.} 
In particular, actor–critic algorithms such as DDPG and TD3 gain substantial improvements from \textit{ReaCritic}'s reasoning mechanism. \textcolor{black}{The dual-axis reasoning architecture stabilizes Q-value estimation and effectively mitigates value degradation, a critical bottleneck in the long-horizon and highly coupled environments characteristic of large-scale wireless networks.} 
Moreover, methods such as A3C and PPO, though less dependent on accurate value estimates, also achieve higher returns when combined with \textit{ReaCritic}. 
\textcolor{black}{This suggests that the increased reasoning capacity provided by \textit{ReaCritic} enhances the critic's ability to provide high-quality baseline signals, thereby improving training stability and exploration efficiency even in policy-gradient-based algorithms.} \textcolor{black}{The consistent gains across these diverse strong baselines reinforce the credibility of our approach for deployment in heterogeneous and dynamic wireless communication scenarios.}

\begin{figure*}
\centering
\subfigure[{\textit{MountainCarContinuous-v0}} with SAC.]
{
\includegraphics[width=0.3\textwidth, height=0.18\textwidth]{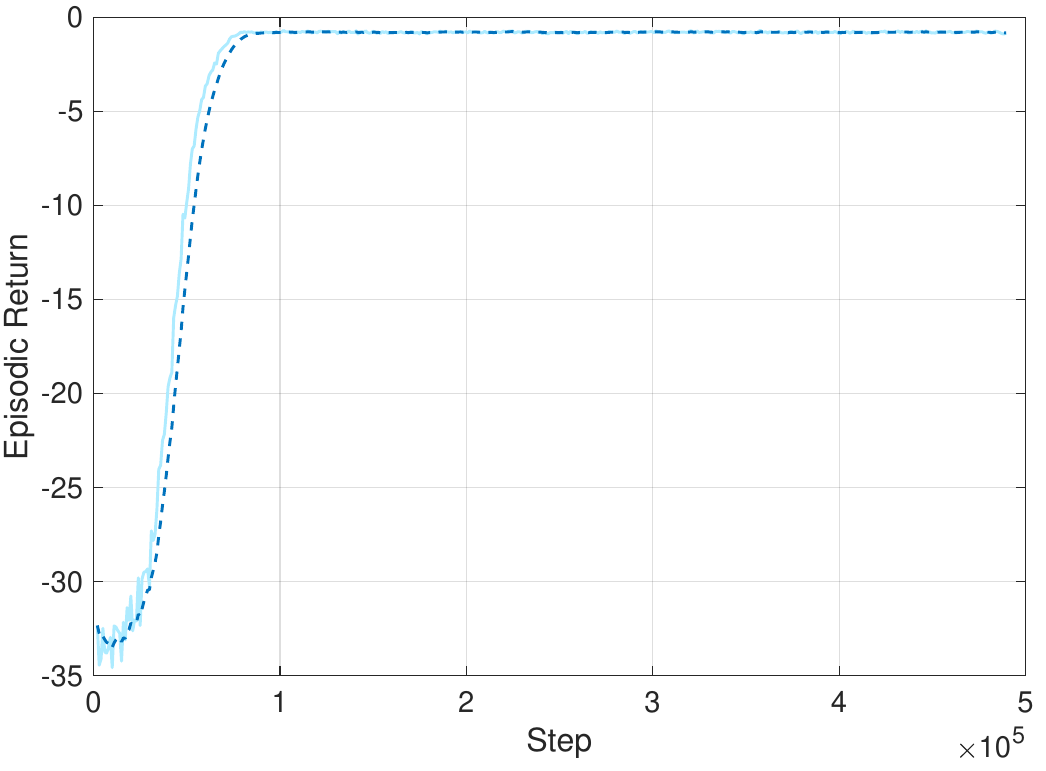}
}
\subfigure[{\textit{HumanoidStandup-v4}} with SAC.]
{
\includegraphics[width=0.3\textwidth, height=0.18\textwidth]{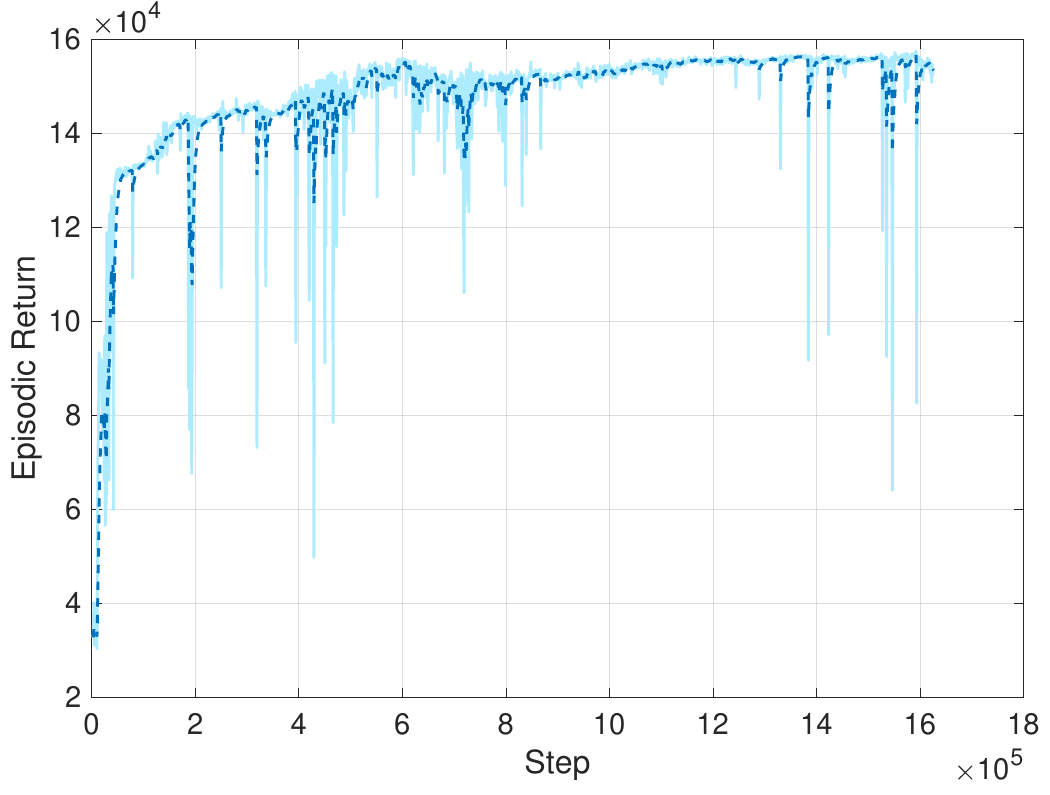}
}
\subfigure[{\textit{Ant-v4}} with SAC.]
{
\includegraphics[width=0.3\textwidth, height=0.18\textwidth]{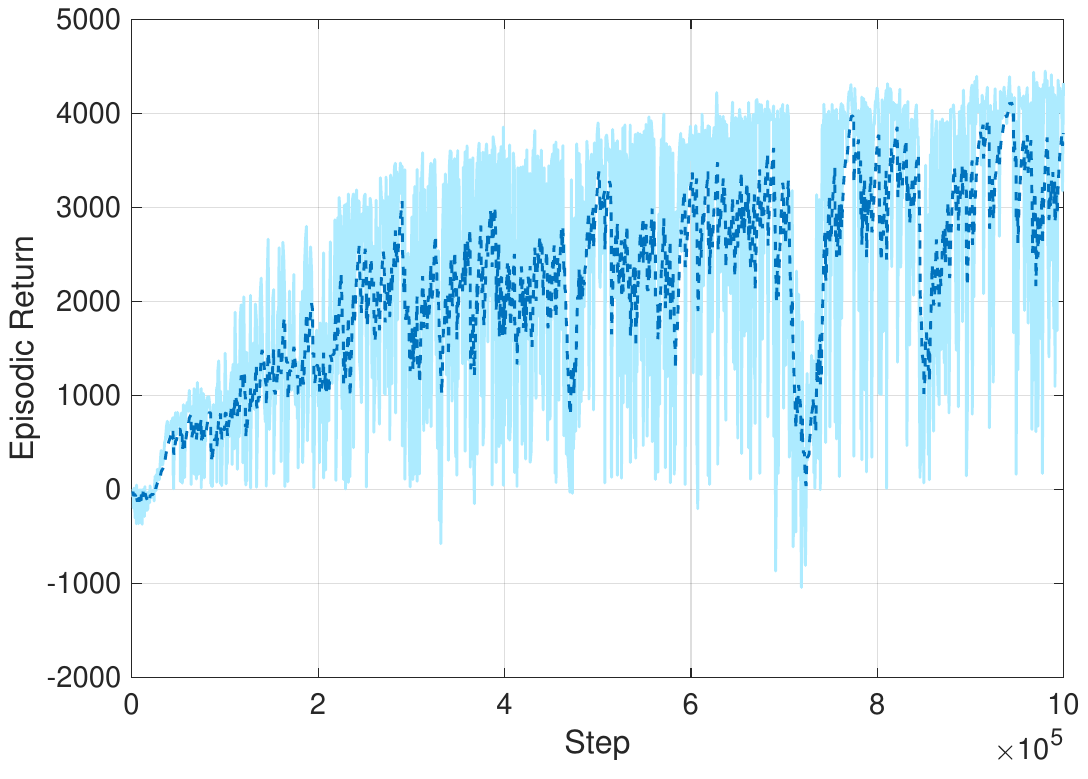}
}
\subfigure[{\textit{MountainCarContinuous-v0}} with {\textit{ReaCritic}}-based SAC.]
{
\includegraphics[width=0.3\textwidth, height=0.18\textwidth]{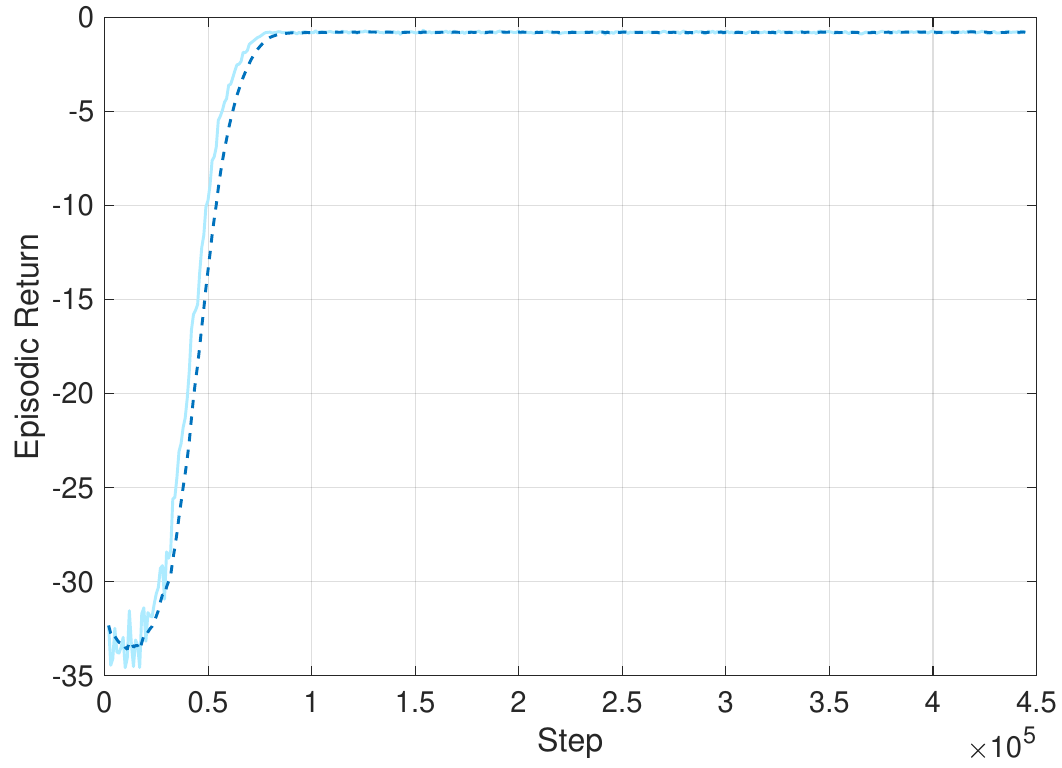}
}
\subfigure[{\textit{HumanoidStandup-v4}} with {\textit{ReaCritic}}-based SAC.]
{
\includegraphics[width=0.3\textwidth, height=0.18\textwidth]{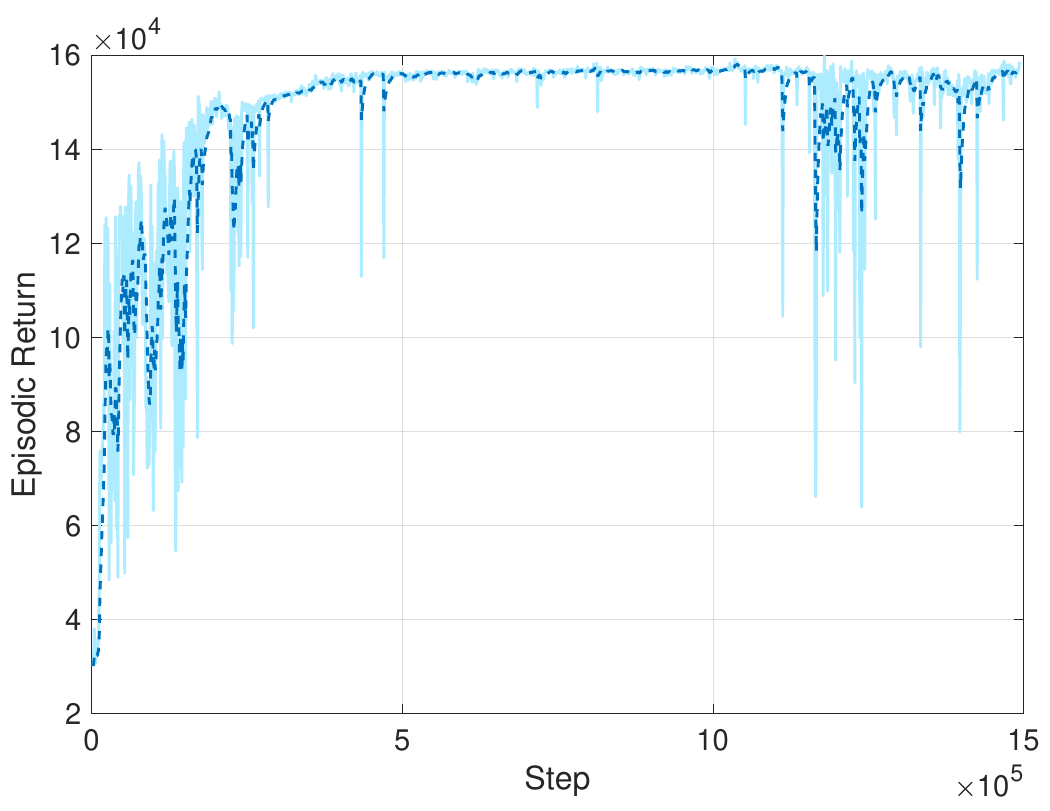}
}
\subfigure[{\textit{Ant-v4}} with {\textit{ReaCritic}}-based SAC.]
{
\includegraphics[width=0.3\textwidth, height=0.18\textwidth]{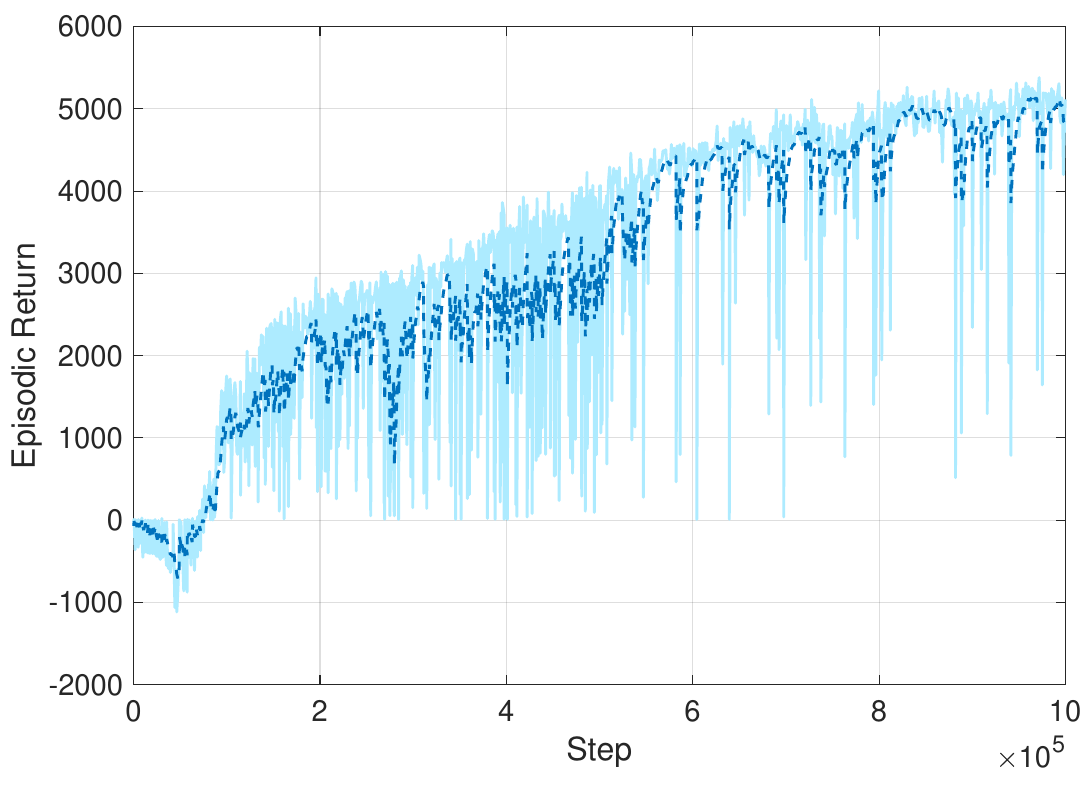}
}

\caption{The comparison of the training performance with  {\textit{ReaCritic}}-based SAC method, and standard SAC under {\textit{MountainCarContinuous-v0}}, {\textit{HumanoidStandup-v4}} and {\textit{Ant-v4}}.}
\label{fig:env}
\end{figure*} 

Fig.~\ref{fig:env} shows that the {\textit{ReaCritic}}-based SAC performs robustly across different environments, including
standard GYM benchmarks. And it is observed that the more complex the environment is, the better {\textit{ReaCritic}}-based SAC outperforms SAC. In the simple environment type such as {\textit{MountainCarContinuous-v0}}, the performance of {\textit{ReaCritic}}-based SAC and SAC are almost the same. In moderately complex environment such as {\textit{HumanoidStandup-v4}} and in high-complexity humanoid control environment
{\textit{Ant-v4}}, the {\textit{ReaCritic}}-based SAC outperforms SAC and achieves higher rewards. And compare Fig.~\ref{fig:env} (b) and Fig.~\ref{fig:env} (e) with the results in Table.~\ref{tab:rea_vs_baseline}, it shows that the proposed {\textit{ReaCritic}}-based SAC algorithms achieve higher performance and more stable convergence than the other {\textit{ReaCritic}}-based DRL algorithms and DRL benchmark algorithms.

\subsubsection{Ablation Study on Horizontal Noise Exploration in HRea}
\begin{figure}
\centering
{
\includegraphics[width=0.4\textwidth]{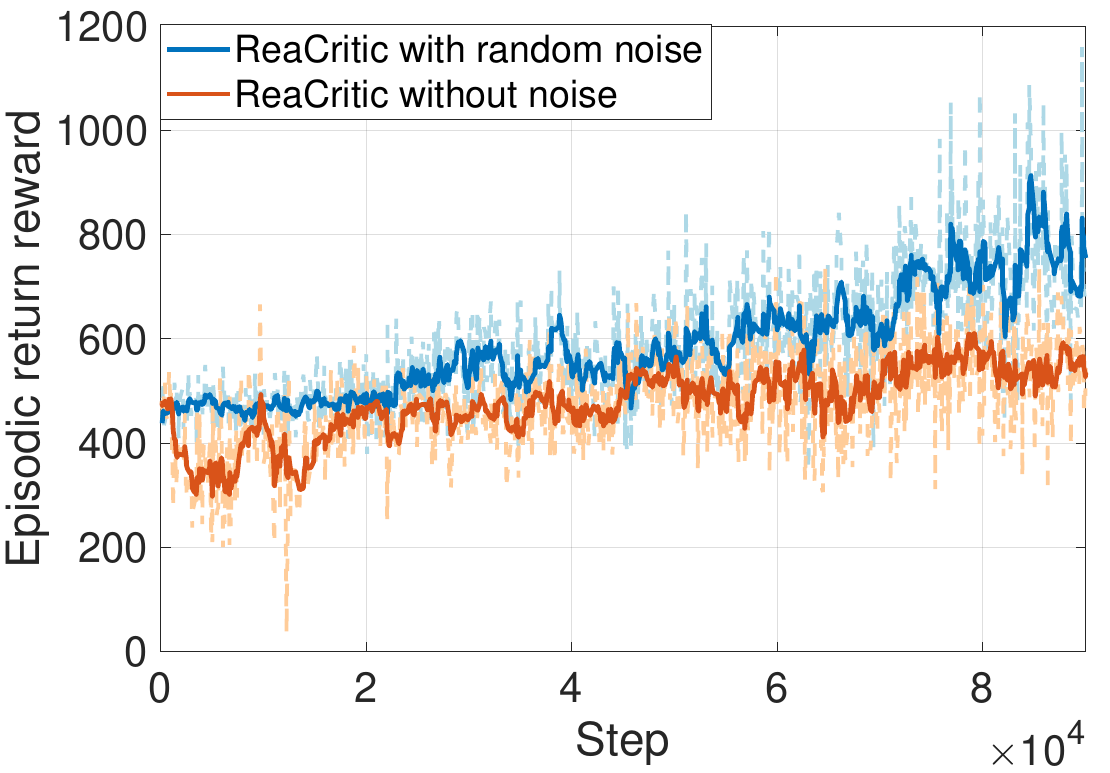}
}
\caption{Reward performance for {\textit{ReaCritic}}-based SAC with random noise and without noise when user number is  $50$.}
\label{noise}
\end{figure}

We evaluate the effect of horizontal token perturbation in HRea by removing the noise injection and comparing its impact on learning stability and final performance. One key design in \textit{ReaCritic} is the addition of Gaussian noise to the horizontally expanded tokens to enhance generalization. To assess its utility, Fig.~\ref{noise} presents the episode return of \textit{ReaCritic}-based SAC with and without noise under a 50-user HetNet setting. Results show that noise injection consistently improves training stability and final reward, especially in large-scale networks where modeling diversity and reducing overfitting become more critical.

\section{Conclusion}\label{conclusion}
We proposed \textit{ReaCritic}, a reasoning transformer-based critic-model scaling framework that introduces reasoning-like capabilities inspired by LLMs into DRL for intelligent HetNet management. By combining HRea over diverse state-action sequences with VRea through stacked transformer blocks, \textit{ReaCritic} overcomes the limitations of conventional shallow critics in complex, dynamic environments. Its modular design supports integration with various critic-based DRL algorithms, enabling flexible deployment across tasks. Extensive experiments demonstrate that \textit{ReaCritic} enhances generalization, robustness, and convergence, offering a promising direction for intelligent and scalable network management.

\bibliographystyle{IEEEtran}
\bibliography{Ref}
\end{document}